\documentclass[sigplan,10pt]{acmart}
\renewcommand\footnotetextcopyrightpermission[1]{}
\usepackage{amsmath}
\usepackage{booktabs}
\usepackage{enumitem}
\usepackage{graphicx}
\usepackage{subfigure}
\usepackage{xspace}
\usepackage{xcolor}
\usepackage{algorithm}
\usepackage{algpseudocode}
\usepackage{multirow}
\usepackage{pifont}
\usepackage{listings}

\definecolor{lstkw}{RGB}{0,0,180}
\definecolor{lstcm}{RGB}{0,128,0}
\definecolor{lststr}{RGB}{163,21,21}
\definecolor{lstln}{RGB}{128,128,128}
\definecolor{lstplc}{RGB}{200,30,30}
\lstdefinestyle{embodirl}{
  basicstyle=\ttfamily\footnotesize,
  keywordstyle=[1]\color{lstkw}\bfseries,
  keywordstyle=[2]\color{lstplc}\bfseries,
  commentstyle=\color{lstcm}\itshape,
  stringstyle=\color{lststr},
  showstringspaces=false,
  tabsize=2,
  breaklines=true,
  numbers=left,
  numberstyle=\tiny\color{lstln},
  numbersep=6pt,
  xleftmargin=14pt,
  columns=fullflexible,
  upquote=true,
  morekeywords=[1]{True,False,None,self,num_gpus,mps,cpu_pool,
    stage,sm,mem,worker,cores,src,dst,n,gpus,roles,mode,profile,
    pool,name,model,env,sm_grid,batch_grid,stats,num_iters,it},
  morekeywords=[2]{ResourcePool,EmbodiPlacement,
    HybridComponentPlacement,set_quota,bind,transfer,release,query,
    get_strategy,placement,placement_strategy,create_group,launch,
    Cluster,Worker,WorkerCls,Runner,actor,rollout,
    apply,balance,step,reallocate},
}

\newcommand{\name}{EBRL\xspace}

\newcommand{\eg}{\textit{e.g.}\xspace}
\title{\name: Asynchronous Embodied RL by Multi-Grained Resource Management}

\author{
{Liang Mi$^{1}$\footnotemark[1]\footnotemark[2]\;Weijun Wang$^2$\footnotemark[1]\;
Bowen Gao$^{1}$\;
Tianze Yu$^{1}$\;
Zixu Hao$^{2}$\;
Han Xiao$^{3}$\;
Xin Ding$^{4}$\;
Mingzhe Huang$^{2}$\;
Xin He$^{3}$\;
Lu Shi$^{2}$\;
Hao Wu$^{1}$\;
Haipeng Dai$^1$\footnotemark[3]\;
Guihai Chen$^{1}$\;
Yunxin Liu$^{2}$}
Ting Cao$^{2}$\footnotemark[3]\;\\
$^1$ Nanjing University
$^2$Institute for AI Industry Research (AIR), Tsinghua University
$^3$Nanjing University of Posts and Telecommunications
$^4$University of Science and Technology of China
}
\renewcommand\footnotetextcopyrightpermission[1]{}
\setcopyright{none}
\acmDOI{}
\acmISBN{}
\acmPrice{}

\begin{document}

\begin{abstract}
Embodied reinforcement learning (RL) improves model capabilities with a pipeline of environment simulation, action generation, and model updates.
These stages show heterogeneous CPU and GPU demands, making efficient resource utilization difficult.
Recent systems overlap rollout (simulation and generation) with training for efficiency, but exclusive GPU allocation and synchronized barrier in rollout still leave substantial hardware resource waste.
In this paper, we present \name, an asynchronous embodied RL training system with two core techniques.
The asynchronous pipelined scheduler overlaps rollout and training, pipelines simulation and generation across environment groups, and carries out each environment independently, eliminating synchronization stalls.
The fine-grained resource manager pools CPU cores and GPU streaming multiprocessors, and uses stage profiles and runtime feedback to adjust resource quotas and batch sizes to meet the shifting demands among stages.
We implement \name on RLinf and evaluate it with four embodied policies and four simulation benchmarks across heterogeneous GPU testbeds.
Experiments show that \name achieves $1.30\times$--$3.47\times$ the end-to-end rollout throughput and $2.5\times$ time of training convergency compared to the SOTA embodied RL systems.
% It also completes a LIBERO-Object/GR00T training run approximately $2.5\times$ faster than synchronous execution while maintaining a comparable final success rate.
\end{abstract}

\maketitle

\renewcommand{\thefootnote}{\fnsymbol{footnote}} %将脚注符号设置为fnsymbol类型，即特殊符号表示

% \footnotetext{This paper has been accepted by NSDI.}
\footnotetext[1]{Liang Mi and Weijun Wang contributed equally to this work.} %对应脚注[1]
\footnotetext[2]{This work was done while Liang Mi interns at the Institute for AI Industry
Research (AIR), Tsinghua University.} %对应脚注[2]
\footnotetext[3]{Corresponding author: Ting Cao and Haipeng Dai.}

\section{Introduction}

% 应用背景->RL post-training成为de facto->三阶段异构流水线->现有系统->三个limitation->root cause + research question->两个challenge->我们的系统->挑战-技术一一对应->贡献
% 篇幅对照KVCodec: P1~90 P2~115 P3~95 P4~105 P5~120 C1~95 C2~130 P8~25 T1~95 T2~85 Contrib~110
Embodied models have become a foundational building block for physical intelligence.
By mapping multimodal observations, including visual inputs, proprioceptive feedback, and language instructions, to continuous control actions, they power diverse embodied tasks, from dexterous manipulation~\cite{dexmanip_openai, dexmanip_demo, mobilealoha} and legged locomotion~\cite{locomotion_wild, quadruped_terrain, h2o} to household navigation~\cite{habitat, octo} and long-horizon planning~\cite{saycan}.
% \weijun{It doesn't make sense to expand on the architecture of VLA and WAM here.} 
% As these policies scale in capability, notably through vision-language-action (VLA) models~\cite{rt2} and world action models (WAMs)~\cite{dreamzero}, reinforcement learning (RL) post-training has become the de facto approach for adapting them to specific deployment tasks~\cite{vla2rl, dppo}.

\begin{figure}[t]
\centering
\includegraphics[width=1\linewidth]{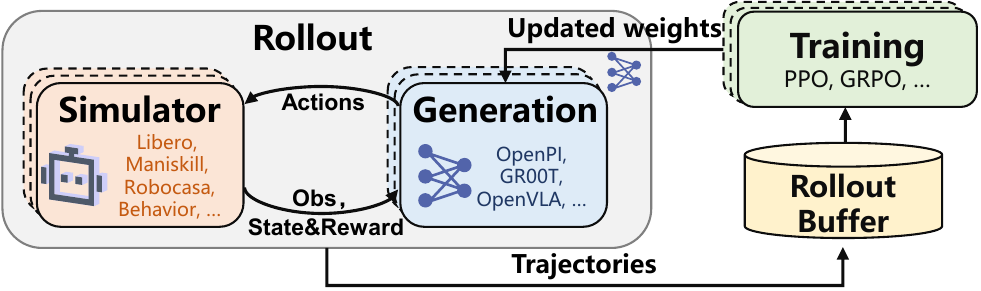}
    \caption{Overview of the embodied model training pipeline.}
    \label{fig:RL_pipeline}
    \vspace{-1.5em}
\end{figure}

To boost embodied model ability, reinforcement learning is widely used in physical intelligence today. %adopted approach 
It places the models into target environments and improves their task-completion capacity through trial and error~\cite{vla2rl, dppo, tang2025deep,lu2025vla,kalashnikov2018scalable, intelligence2025pi,hu2025flare,huang2026thinkact,zou2026d2ppo,kaufmann2023champion,tan2025interactive}.
% by enabling them to adapt to specific deployment tasks.
% However, RL post-training introduces a fundamentally different execution structure from supervised fine-tuning.
% Because distribution shifts in observations, action spaces, and rewards prevent pretrained policies from transferring directly to deployment tasks, RL post-training adapts them by enabling closed-loop interaction with a simulator, where the policy improves iteratively under reward signals.
% \weijun{The logic doesn't flow. Heterogeneity occurs suddenly. This paragraph should be an introduction to embodied RL pipeline and a summary of heterogeneity at the end.}
As shown in Fig.~\ref{fig:RL_pipeline},
embodied RL training comprises two stages, rollout and training.
% \textit{simulation}, \textit{generation}, and \textit{training}.
In each iteration, the \textit{simulator} interacts with model \textit{generation} to roll out enough trajectories (\eg, 2K pieces), and then the \textit{training} stage ingests them to update model weights for the next iterative rollout.
Studies~\cite{rlinfvla, simplevlarl, vla2rl, zhai2025vision} from the AI community show that RL can increase the success rate of embodied models by up to more than 40\%.
% with the first two forming the rollout.

Despite these benefits, the complexity of embodied RL pipelines introduces many system challenges.
% Despite these benefits, efficiently utilizing hardware resources for embodied RL remains challenging.
Its multiple stages exhibit highly heterogeneous computational patterns.
The simulator requires CPU-GPU co-processing to simulate physical interactions and render visual observations, while generation and training are typical matrix computations that rely heavily on GPUs (\S\ref{sec:rl_for_embodied}).
%cpu-simu，gpu-render
Furthermore, to accelerate trajectory collection, rollout typically simulates hundreds of environments concurrently and generates actions in batches.
This large-scale parallelism further complicates resource management and scheduling.
% This massive parallelism significantly exacerbates resource contention and scheduling complexity.
% Simulator 计算robot动作和环境的交互并渲染环境画面变化，通常需要CPU和GPU共同参与; while generation 和 training是典型的模型推理和训练计算rely on GPU。(more in)
% In addition, 为了快速收集轨迹提升模型训推效率, simulator通常并发上百个环境，这进一步加剧了资源的难度。

% 和runs the model on GPUs to process each observation and emit the next action chunk

% Such a  incurs very complicated resource management.

% exclusive GPU allocation
% \textcolor{red}{the simulator advances hundreds of parallel environments by performing physics simulation and sensor rendering, largely on CPUs; generation runs the model on GPUs to process each observation and emit the next action chunk; and the two interleave turn by turn until complete trajectories emerge.
% Training then performs gradient updates on GPUs over the collected trajectories.
% This three-stage loop constitutes a heterogeneous CPU-GPU pipeline whose training efficiency hinges on how resources are split across stages (\S\ref{sec:rl_for_embodied}).}

\begin{figure}[t]
\centering
% \begin{minipage}[b]{1\linewidth}
    % \centering
    \includegraphics[width=0.85\linewidth]{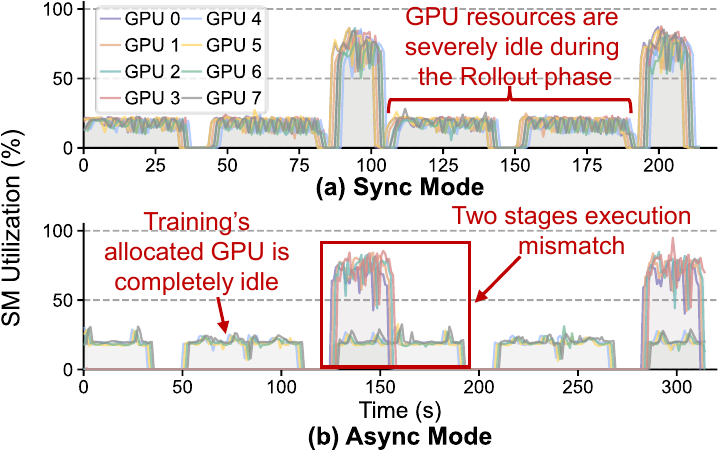}
    \caption{Severe resource waste when training OpenPI~\cite{pi0} model on RoboCasa~\cite{robocasa} simulator with RLinf.}
    \label{fig:utl_overall}
% \end{minipage}
\vspace{-1.5em}
\end{figure}

Modern embodied RL systems mitigate these issues with flexible resource allocation and asynchronous execution.
% learning inefficiency替换，后面介绍同步异步如何缓解
% \weijun{The following sentence delivers zero information. "sync system execute stages syncly", so wiered.}
They dynamically adjust resource allocation across stages~\cite{rlinf,rlinfvla,rlinf_user, 318473} and asynchronously overlap rollout with training to mask long-tail latency\footnote{One slow environment (\eg, the failed trajectories roll out until maximum step) can delay the whole rollout stage due to the synchronized barrier of batching generation.} (\S\ref{sec:limitations})~\cite{streamrl, areal, asyncrlhf, dvla, rlvla3}.
% while, asynchronous RL pipeline further improves the resource utilization by
%灵活的资源分配
% For futher ,
However, these systems suffers from coarse-grained resource management.
For example, the exclusive GPU allocation for generation and training stages causes severe waste.
As traced in Fig.~\ref{fig:utl_overall}, over 90\% of the RL time, training OpenPI~\cite{pi0} model on RoboCasa~\cite{robocasa} simulator produces less than 50\% of the GPU utilization.
% regardless of whether the pipeline is synchronous or asynchronous.
%with best efficient mode\footnote{Experiment details are provided in \S\ref{sec:evalutaion}.} 
% Even worse, these underutilization cannot be mitigated by simply increasing the batch size (\S\ref{sec:limitations}) due to the limited throughput of the simulator \weijun{why mentioned this? I don't think reviewer will think about this as no experiment setting clarified here}.
% A naive solution is to increase the batch size to saturate the GPU, but it doesn't work because the limited throughtput of simulator.
% \weijun{No matter how large the batch size is set, gpu cannot be fully utilized?}

To tackle this waste, we ask: \textit{Can we reclaim the idle resources for other stage execution, enabling efficient embodied RL?}
% implement an asynchronous RL system with fine-grained resource management, enabling efficient embodied RL?}
The answer is yes.
Our key design choice is to overlap stages at multiple granularities asynchronously, making interleaved stages reclaim otherwise idle hardware capacity.
However, this poses two challenges.
% Sustaining concurrent execution across dependent stages and allocating resources to match their processing rates as workloads change.

First, sustaining overlap across stages is difficult due to the simulation's synchronous barrier and data dependency during rollout.
% because stages depend on one another and simulation shows a 
On the one hand, the synchronous barrier of simulation incurs a long-tailed issue.
Heterogeneous simulation paces across environments create stragglers, forcing faster environments to stall at batch synchronization points and leading to higher end-to-end latency and wasted compute resources.
On the other hand, even with an asynchronous rollout and training, as in prior systems, data dependency between simulation and generation still causes pipeline bubbles (\S\ref{sec:challenges}).
% simulation and generation interact with actions and observations, while training consumes trajectories and publishes updated model weights.
% Overlapping rollout and training alone still leaves simulation and generation alternating within each rollout batch.
% Simulation latency variance across environments further creates waiting at batch barriers.
% Futhermore, the simulator's long-tailed latency can cause some environments to finish much later than others, forcing all other environments to wait at the batch barrier .

Second, to reclaim idle hardware, the system must partition GPUs and CPUs into sub-device units and assign each stage appropriate resources.
% Although the modern hardware provide sharing mechanisms~\cite{cuda_mps,nvidia_mig}, deriving the right split
This is non-trivial.
The optimal allocation depends on complex, coupled interactions among hardware architectures, models, task complexity, and training parameters (\eg, the number of environments), making it impractical to comprehensively evaluate all possible allocation strategies.
% a search space infeasible to profile exhaustively.
More difficult, the optimal allocation changes as training progresses.
% stage costs drift 
Along with model updates, success trajectories terminate earlier, so resource demand across stages shifts.
A static strategy inevitably misallocates resources and bottlenecks the pipeline (\S\ref{sec:challenges}).

In this paper, we present \name, an asynchronous embodied RL training system powered by fine-grained resource management, to address the challenges outlined above.

We address the first challenge with \textit{asynchronous pipelined scheduling} (\S\ref{sec:design-T2}).
Stage-level async-pipelined execution (\S\ref{sec:design-inter} \&  \S\ref{sec:design-intra}) overlaps rollout and training across iterations and pipelines simulation with generation across environment groups.
Generation and training execute concurrently on shared GPUs under separate logical SM quotas.
Step-level asynchronous orchestration (\S\ref{sec:design-step}) allows environments to advance independently and rebinds freed CPU cores to other stragglers, reducing waiting due to environment variability.

To address the second challenge and implement multi-grained asynchronous execution, we propose \textit{fine-grained resource management} (\S\ref{sec:design-T1}).
The \textit{GPU-CPU resource pool} (\S\ref{sec:design-pool}) exposes CPU cores and GPU SMs as sub-device schedulable resources.
The \textit{elastic resource allocator} (\S\ref{sec:design-alloc}) uses offline profiles to jointly select per-stage resource quotas and batch sizes for end-to-end throughput.
At runtime, it monitors stage throughputs and adjusts allocations when a sustained imbalance is detected.

We summarize our key contributions as follows:

\scalebox{0.8}{$\bullet$} We characterize hardware underutilization in embodied RL training and show that it stems from coarse-grained, stage-exclusive resource allocation and synchronization barriers across rollout iterations.

\scalebox{0.8}{$\bullet$} We prototype \name, an efficient embodied RL training system.
It involves two core techniques, asynchronous pipelined scheduler and fine-grained resource management, exposed through declarative programming abstractions.

\scalebox{0.8}{$\bullet$} We implement \name and conduct evaluations on four embodied models covering both VLA and WAM families, four CPU- and GPU-accelerated simulators, and three heterogeneous testbeds.
Experimental results show that \name achieves up to $1.98-2.98\times$ throughput improvement over SOTA systems.

% \textbf{This work does not raise any ethical issues.}

\section{Background}

\subsection{Embodied Models}
Embodied models convert multimodal observations into robot actions.
Current architectures fall into two families: \emph{Vision-language-action (VLA) models} and \emph{World action model (WAM) policies}.
They share the same input interface but differ in how that interface is turned into actions.
Both ingest visual observations, robot state (\eg, joint angle and velocity), and natural-language task instruction, and generate an \emph{action chunk} (\eg, next 5 or 8 actions) by iteratively denoising a random tensor over multiple steps.
% produced by iteratively denoising a randomly initialized tensor over multiple denoising steps.
VLA~\cite{rt2, openvla, pi0, groot, hpt, vlasurvey} delegate this denoising to a separate diffusion-based action head. 
The vision-language-model (VLM) backbone takes all inputs, fuses them into a shared representation (\eg, KV cache, special tokens), and feed it into the action head to yield the action chunk.
% ingest the visual observations, the robot states (\eg, joint angle and speed), and the task instruction, then output the next action, as shown on the left of Fig.~\ref{fig:model_arch}.
% Specifically, the VLM backbone encodes the visual observations into visual tokens and embeds the task instruction and robot state into text tokens, which are then fused together through the LLM backbone into a shared representation (\eg, KV cache, special tokens).
% This representation is then fed into a diffusion-based action head, which denoises a randomly initialized action tensor over multiple denoising steps to yield the final action chunk.
Differ from that, WAM~\cite{dreamzero} adopt the Diffusion Transformer (DiT)~\cite{dit} as the backbone.
% Similar to VLA models, the DiT encodes the visual, language, and proprioceptive inputs into a shared representation. 
It cosumes all inputs and jointly denoises a sequence of latent future frames together with the action chunk, then per-modality decoders seperately generate the predicted frames and action chunks.
WAM's action generation is anchored to the model's anticipated visual outcome rather than produced by a downstream head, so it often achieve better performance on long-horizon tasks.                                                    
% However, instead of feeding it into a separate action head, the DiT itself first predicts future visual states (\eg, using a visual decoder to generate predicted frames), uses these predictions to guide action generation, and then denoises the final action chunk, as shown .

\subsection{RL for Embodied Model}
\label{sec:rl_for_embodied}

Reinforcement learning has been proven to be an efficient way to enhance embodied model
%  on target deployment tasks
~\cite{vla2rl, dppo}.
% Unlike supervised fine-tuning, which consumes a static offline dataset, 
It updates the policy in a closed loop with a simulator, alternating between collecting trajectories (training data) and updating parameters on them. 
% Because the policy keeps changing across iterations, fresh trajectories must be regenerated every iteration, which turns simulation into a dedicated compute stage that runs continuously alongside policy model inference and gradient updates.
In particular, each training iteration includes three stages, \textit{simulation}, \textit{generation}, and \textit{training}, with distinct compute patterns, as shown in Fig.~\ref{fig:RL_pipeline}; and simulation and generation forms rollout.
% The policy interacts with a simulator in closed-loop fashion, collecting trajectories and updating parameters in alternation.
% Unlike offline datasets, the simulator generates new trajectories on demand for each training iteration.
% This turns simulation into a dedicated compute stage that must run continuously alongside policy inference and gradient updates.
% As shown in Fig.~\ref{fig:RL_pipeline}, a single iteration therefore decomposes into three stages with distinct compute patterns:

\noindent\textbf{Simulator}, as depicted in Fig.~\ref{fig:RL_pipeline}, computes physical interaction and renders observations interacting with the embodied model.
A simulator stack is organized into three layers, as summarized in Table~\ref{tab:sim_platforms}.
At the bottom, an \emph{engine} performs the actual computation, splitting into a \emph{physics engine} that solves rigid-body dynamics, contact resolution, and collision detection, and a \emph{rendering engine} that synthesizes the visual observations. 
% the two are decoupled and may run on different hardware.
On top of an engine, a \emph{framework} wraps it with robot models and environment APIs (\eg,  RoboSuite~\cite{robosuite} over MuJoCo~\cite{mujoco} in Table~\ref{tab:sim_platforms})
% , SAPIEN~\cite{sapien} over PhysX~\cite{physx}, and Isaac Sim/Omniverse~\cite{isaaclab} over PhysX with RTX ray tracing.
At the top, a \emph{benchmark} defines concrete tasks, scenes, and assets on a framework (\eg, LIBERO~\cite{libero} and RoboCasa~\cite{robocasa} on RoboSuite).
% , ManiSkill~\cite{maniskill} on SAPIEN, and Isaac Lab~\cite{isaaclab} and BEHAVIOR~\cite{behavior} on Isaac Sim.
The choice of simulator stack dictates the hardware footprint: MuJoCo-based benchmarks are CPU-only, whereas PhysX-based benchmarks are CPU-GPU (with RTX stacks further requiring RT cores).

\begin{table}[t]
\centering
\small
\setlength{\tabcolsep}{4pt}
\caption{Representative embodied benchmarks decompose into a framework and an underlying physics/rendering engine, yielding distinct hardware footprints.}
\label{tab:sim_platforms}
\begin{tabular}{lll}
\toprule
Benchmark & Framework & Physics / Rendering \\
\midrule
LIBERO~\cite{libero} & \multirow{2}{*}{RoboSuite} & \multirow{2}{*}{MuJoCo(CPU) / OpenGL(GPU)} \\
RoboCasa~\cite{robocasa} & &\\
\midrule
ManiSkill~\cite{maniskill} & SAPIEN & PhysX(CPU/GPU) / Vulkan(GPU) \\
\midrule
Isaac Lab~\cite{isaaclab} & \multirow{2}{*}{Isaac Sim} & \multirow{2}{*}{PhysX(CPU/GPU) / RTX(GPU)} \\
BEHAVIOR~\cite{behavior} & & \\
\bottomrule
\end{tabular}
\end{table}

\noindent\textbf{Generation and training} performs and updates the embodied model to generate actions interacting with simulator.
% in contrast to the simulator's mixed CPU-GPU footprint, run entirely on GPUs.
% entires execute on GPUs cover the remaining two stages of Fig.~\ref{fig:RL_pipeline}. 
Generation, in each rollout step,
% , interacting with simulator step-by-step during rollout, 
ingests the observations, states, and rewards from the simulator, then produces the next action chunk. 
% and save the collected trajectories into a replay buffer for training.
Training updates the policy using trajectories collected during rollout with RL algorithms (\eg, PPO~\cite{ppo} or GRPO~\cite{grpo}), which are then used by generation for the next iteration of rollout.
The rollout trajectories are stored in a buffer for training use.
% However, the GPU compute they demand varies widely with the different policy models and the training method choices.
% % Unlike the simulator's mixed CPU--GPU footprint, both stages run entirely on GPUs, yet the GPU compute they demand varies widely with the policy along two axes.
% \textcolor{red}{First, embodied policy models span a large capacity range, from \tododata{$\sim$300M}-parameter lightweight models~\cite{octo, diffusionpolicy} to \tododata{$\sim$7B}-parameter foundation policies~\cite{openvla, pi0, groot}.
% Second, the model update regime differs across model components: the vision encoder is typically frozen and runs forward-only, the backbone is updated via either LoRA adapters or full-parameter fine-tuning when compute permits, and the action head (or the action expert in $\pi_0$-style designs) is always fully fine-tuned~\cite{openvla, pi0}.}
% As a result, the GPU footprint of generation and training shifts substantially across model choices and training configurations.

\begin{figure}[t]
\centering
\includegraphics[width=1\linewidth]{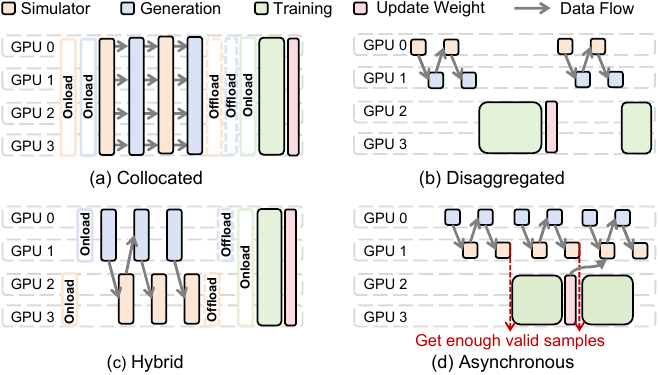}
    \caption{Execution modes for embodied RL systems.}
    \label{fig:exec_mode}
\end{figure}

\subsection{Embodied Reinforcement Learning Systems}
To support the heterogeneous workload of three stages for efficient embodied RL training, modern systems~\cite{rlinf, rlinfvla, rlinf_user, dvla, rlvla3} propose four execution modes to orchestrates GPU cluster, as shown in Fig.~\ref{fig:exec_mode}.
\textit{Colocated} mode shares each GPU in temporal across all three stages and every stage runs in turn on every GPU. 
However, it causes GPU idleness during stage switching, making the GPU underutilized.
%to host memory to free GPU memory for the active stage, eliminating cross-stage idleness at the cost of repeated state offloading.   
\textit{Disaggregated} mode partitions GPUs in spatial across stages, in which each stage owns their allocated and non-overlapping GPU.
While this eliminates the context-switching overhead of the colocated mode, it introduces severe waiting times between rollout and training.
To reduce this waiting time, \textit{Hybrid} mode orchestrates GPUs in both temporal and spatial.
During rollout, simulator and generation occupy separate GPUs and pipeline execution; while to computation-heavy training stage, all GPUs are grouped together for gradient computation over the collected rollout data.     
Notably, no single above mode dominates.
Colocated tends to win for small models where onload/offload overhead is low, while disaggregated tends to win for large ones.
% \weijun{For now, asynchronous becomes a key feature, which occurs in the paper title, it should be placed in the method.}
% \textcolor{red}{
\textit{Asynchronous} mode, first developed in LLM RL systems~\cite{streamrl, areal, asyncrlhf} and recently adapted to embodied RL~\cite{dvla, rlvla3}, removes the stage barrier between rollout and training by overlapping them on separate GPU pools with off-policy correction algorithms that tolerate moderate policy staleness.

\section{Motivation and Challenges}
This section analyzes the limitations of existing embodied RL systems and identifies two fundamental challenges that constrain their efficiency.

\subsection{Limitations of SOTA Systems}
\label{sec:limitations}
% Despite delivering real speedups, existing embodied RL systems allocate resources at whole-GPU granularity and treat CPU cores as unmanaged background capacity.
To analyze the limitations of SOTA embodied RL systems in depth, we benchmark the training of OpenPI~\cite{pi0} on RoboCasa~\cite{robocasa} using PPO.
Our testbed pairs Intel Xeon Gold 6348 CPUs (56 cores) with 8$\times$NVIDIA A100 GPUs.
We evaluate two representative mode, colocated mode in RLinf~\cite{rlinf} and asynchronous mode in D-VLA~\cite{dvla}, which represent the SOTA in synchronous and asynchronous execution, respectively.
% RLinf~\cite{rlinf} and D-VLA~\cite{dvla}, which represent the state-of-the-art in synchronous and asynchronous execution, respectively.
% The entire training procedure relies on RLinf in collocated mode, as it delivers the best performance among the options.
% With deep analysis, we observe three limitations.

\begin{figure}[t]
\setlength{\abovecaptionskip}{3pt}
\centering
\begin{minipage}[b]{0.49\linewidth}
    \centering
\includegraphics[width=1\linewidth]{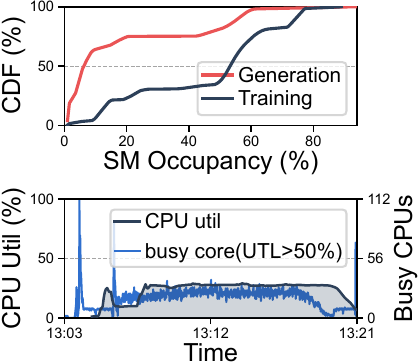}
\caption{SM and CPU occupancy during RL training. 
% \weijun{The font is too small, and the legend is 50 80? what is that?}
}
    \label{fig:sm_cpu_occupancy}
\end{minipage}
\begin{minipage}[b]{0.49\linewidth}
\centering
\includegraphics[width=1\linewidth]{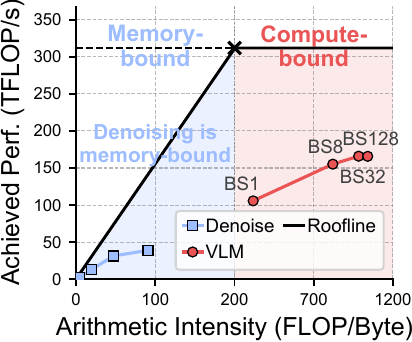}
\caption{Roofline of VLA rollout on one A100.}
\label{fig:vla_roofline}
\end{minipage}
\vspace{-1em}
\end{figure}

% \begin{figure}[t]
% \setlength{\abovecaptionskip}{3pt}
% \centering
% \begin{minipage}[b]{0.49\linewidth}
%     \centering
% \includegraphics[width=1\linewidth]{figs/vla_roofline.pdf}
%     \caption{Roofline of VLA rollout on one A100.}
%     \label{fig:vla_roofline}
% \end{minipage}
% \begin{minipage}[b]{0.49\linewidth}
% \centering
% \includegraphics[width=1\linewidth]{figs/robocasa_throughput_vs_batchsize.pdf}
% \caption{Throughput with increasing batch sizes.}
% \label{fig:batchsize_throughput}
% \end{minipage}
% \vspace{-1em}
% \end{figure}

% \begin{figure}[t]
% \setlength{\abovecaptionskip}{3pt}
% \centering
% \includegraphics[width=0.9\linewidth]{figs/vla_roofline.pdf}
%     \caption{Roofline of VLA rollout on one A100 (312~TFLOP/s FP16 compute roof, ridge at $\sim$200~FLOP/Byte). Denoising kernels stay memory-bound with achieved throughput below 40~TFLOP/s, while VLM kernels enter the compute-bound region but plateau near 165~TFLOP/s ($\sim$53\% of the roof) due to inefficient generic operators.}
%     \label{fig:vla_roofline}
% \vspace{-1em}
% \end{figure}

\noindent\textbf{Limitation 1: Severe underutilization of GPU.}
Existing embodied RL systems assign each entire GPU to one stage, causing severe GPU underutilization.
Fig.~\ref{fig:sm_cpu_occupancy} plots the cumulative distribution function (CDF) of GPU SM and CPU occupancy during generation and training.
More than 75\% of generation time operates with SM occupancy below 20\%.
% Even increasing batch size cannot saturate the GPU due to simulator throughput limits and memory constraints (Fig.~\ref{fig:batchsize_throughput}).
This stems from two factors.
(1) Heavy CPU computing blocks GPU execution.
Embodied RL requires intensive CPU operations such as physics simulation, environment resetting, and observation preprocessing.
These CPU-bound operations stall the GPU, leaving SMs idle while waiting for the next batch of inputs.
(2) Lightweight diffusion-based action heads cannot fully utilize GPU SMs.
The diffusion-based action heads perform iterative denoising through multiple lightweight steps.
Fig.~\ref{fig:vla_roofline} shows these operations are memory-bound: each denoising step launches small kernels, leaving SMs heavily underutilized.

\noindent\textbf{Limitation 2: Lack of CPU management.}
Existing embodied RL systems lack explicit CPU resource management, leading to severe resource contention and context-switching overhead across simulation environments.
Fig.~\ref{fig:sm_cpu_occupancy} plots the CPU utilization per core during rollout: only 22 of the 56 cores sustain utilization above 50\% and very few cores can be saturated.
%  \weijun{seems cannot match figure 4}.
This is because each simulation step requires lightweight operations such as controller processing of actions, setting actuator targets, and post-processing observations.
These operations are frequent yet too small to saturate a CPU core.
%  \weijun{this seems the new reason is this verion paper. double check its correctness}.
Worse, without explicit CPU affinity, the OS scheduler frequently migrates these lightweight threads across cores, incurring context switch overhead that compounds the underutilization.

\noindent\textbf{Limitation 3: Lack of fine-grained asynchronous mechanism.}
Recent systems~\cite{dvla, rlvla3} adopt an asynchronous pipeline to overlap rollout and training for efficiency, but the synchronized barrier between simulation and generation within rollout still wastes resources.
Action generation waits for all environments to complete their simulation steps, and conversely, simulation cannot proceed until all environments receive their generated actions.
% serialize simulation and generation within each rollout step through batch synchronization:
% generation waits for all environments to complete their simulating steps, while all environments start simulating step for all to receive actions.
As shown in Fig.~\ref{fig:utl_overall}, with the SOTA asynchronous pipeline~\cite{dvla}, GPUs still have 47\% idle time during embodied RL.
% This underutilization stems from synchronization between simulation and generation, compounded by differences in execution time between rollout and training.
% This stop-and-go pattern preventsd continuous CPU-GPU overlap within rollout \weijun{is this last sentence right?}.

\begin{figure}[t]
\setlength{\abovecaptionskip}{3pt}
\centering
\begin{minipage}[b]{0.49\linewidth}
    \centering
\includegraphics[width=1\linewidth]{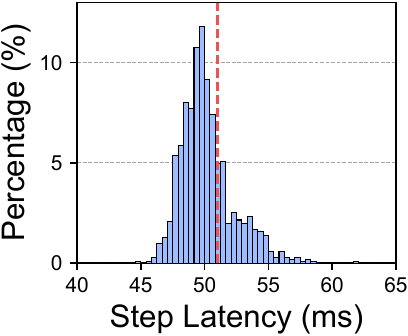}
\caption{Per-step simulator latency shows high variance: the slowest 10\% of steps take $>$1.3$\times$ the median.}
\label{fig:env_distribution}
\end{minipage}
\begin{minipage}[b]{0.49\linewidth}
\centering
\includegraphics[width=1\linewidth]{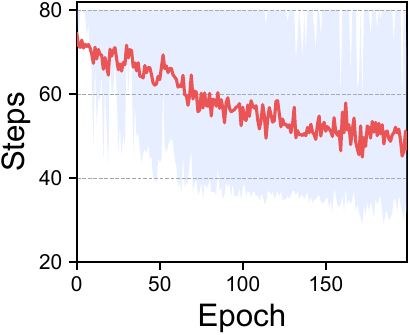}
\caption{Trajectory length decreases as policy improves, making initial resource allocation miscalibrated over time.}
\label{fig:trajectory_length}
\end{minipage}
\vspace{-1em}
\end{figure}

\subsection{Challenges}
\label{sec:challenges}
% The resource underutilization observed in \S\ref{sec:limitations} creates an opportunity for resource sharing.
% When CPU simulation limits the supply of observations, generation cannot fully use its allocated GPU capacity \weijun{where is it?}.
To reclaim the wasted resources, multiple stages must run concurrently by resource sharing.
However, it raises two challenges.
% Since one single stage hardly fully utilize its resources, running multiple stage concurrently on shared resource could reclaim this capacity.
% To realize this opportunity requires sustaining concurrent execution and matching resource quotas to the demands of all three stages.

\noindent\textbf{C1: Inefficient pipeline across heterogeneous stages.}
Sharing GPUs between generation and training stages to overlap rollout with training still leaves GPUs idle.
% requires both stages to execute concurrently. 
% Training updates the model using already collected trajectories while rollout collects new ones.
The reason is as follows. 
% constrained at two finer granularities.
% Running generation and training concurrently on shared GPUs requires overlapping rollout with training on previously collected trajectories \weijun{what do you mean overlap rollout and training, on prevous data? i can't understand}.
% However, this overlap alone still leaves CPU and GPU resources idle for two reasons.
% \weijun{what is the relation of these two reasons?}
% ({\romannumeral 1}) \textit{Synchronization across environments.}
% \weijun{the logic is quite a chaos. what is the logic of next four sentences? please check the logic and rewrite them.}
Existing systems typically execute simulation steps in a synchronized manner across environments.
Each of them progresses at their own pace, after which their final observations are aggregated into a batch and passed to the generation stage for action generation.
However, step latency varies across environments.
As shown in Fig.~\ref{fig:env_distribution}, the average step time is 40~ms, but P99 latency reaches 60~ms.
This variation arises from differences in physical states across environments (\eg, the number and coupling of contact and friction constraints), leading to varying complexity for the physics solver.
As a result, synchronized simulation causes all environments to wait on the struggler for more than half of the simulation time.
% Due to this variance, faster environments must wait for the slowest environment in the same batch to finish before advancing to the next step.
% ({\romannumeral 2}) \textit{Dependency between simulation and generation.}
Furthermore, even if all environments step in the same pace, the data dependency between simulation and generation still makes GPUs idle (\S\ref{sec:limitations}).
This forces us to orchestrate them in a finer-grained manner.

\noindent\textbf{C2: Difficult resource allocation due to the complex, shifting demands across stages.}
Fine-grained allocation requires partitioning GPUs and CPUs into sub-device units to enable flexible stage allocation.
However, determining the optimal resource allocation is non-trivial for two reasons.
({\romannumeral 1}) \textit{The optimal split depends on complex, coupled stage interactions.}
Although hardware primitives like MIG and MPS provide the mechanisms for sub-device partitioning~\cite{cuda_mps, nvidia_mig}, deciding \emph{how} to allocate is hard.
The optimal allocation of SMs and CPU cores is highly sensitive to hardware, policies, task complexities, and RL configurations.
These coupled factors inflate the configuration space, making exhaustive profiling to derive the correct allocation fundamentally infeasible.
({\romannumeral 2}) \textit{Stage costs shift as training progresses, causing static partitions to be obsolete.}
The computational footprint of each stage is not constant. 
As the policy improves, successful trajectories terminate earlier.
As shown in Fig.~\ref{fig:trajectory_length}, the average trajectory length decreases by 31\% from initialization to convergence. 
This continuously alters the compute balance between the rollout and training stages, making the optimal allocation challenging.

\section{\name Design}
\label{sec:design}

\subsection{Overview}
\label{sec:design-overview}
\begin{figure}[t]
\centering
\includegraphics[width=0.95\linewidth]{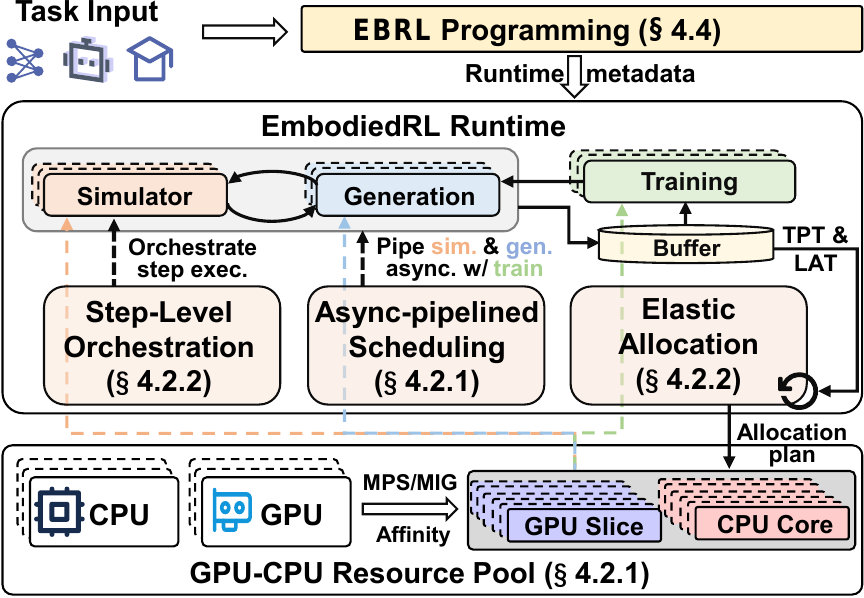}
\caption{\name overview. The runtime coordinates asynchronous execution and elastic allocation over a GPU--CPU resource pool.}
\label{fig:arch}
\end{figure}

% Role: overview -- introduce the architecture modules and their interactions once.
\name coordinates simulation, generation, and training through a GPU--CPU resource pool and three runtime modules, as shown in Fig.~\ref{fig:arch}.
The GPU--CPU resource pool (\S\ref{sec:design-pool}) exposes CPU core sets, GPU SM quotas, and device-memory budgets through a common allocation interface.
Elastic resource allocation (\S\ref{sec:design-alloc}) uses stage performance profiles to select resource quotas and batch sizes, then passes the allocation to the pool.
Asynchronous pipelined scheduler (\S\ref{sec:design-async}) overlaps rollout with training on earlier trajectories and pipelines simulation with generation across environment groups.
Generation and training execute on shared GPUs under the allocated SM quotas.
Step-level orchestration (\S\ref{sec:design-step}) advances environments independently within each action chunk and rebinds freed CPU cores to ready workers.
Runtime throughput measurements feed back to the allocator to adjust stage budgets as training progresses.
The programming interface (\S\ref{sec:design-prog}) accepts the user's policy, simulator, hardware, and training configuration, while the runtime handles resource placement and worker scheduling.

\subsection{Asynchronous Pipelined Scheduler}
\label{sec:design-T2}
\label{sec:design-async}
% Role: opening, challenge, and design -- derive the three levels of overlap.
To use shared CPU and GPU resources effectively, \name must expose concurrent work across the three stages.
However, barriers between iterations, rollout batches, and simulator steps serialize otherwise independent work (\S\ref{sec:challenges}).
We address these barriers at three levels.
Inter-iteration overlap runs rollout alongside training on earlier trajectories.
Within rollout, group-level pipelining overlaps simulation with generation, while step-level orchestration lets individual environments advance independently.
These mechanisms operate within stage resource budgets selected by the allocator in \S\ref{sec:design-alloc}.
\subsubsection{Inter-Iteration Rollout--Training Overlap}
\label{sec:design-inter}
% Role: opening, challenge, and design -- distinguish required dependencies from the global barrier.
Overlapping rollout and training requires decoupling trajectory collection from policy updates across iterations.
In synchronous execution, training waits for rollout to finish, and the next rollout waits for training to finish.
Yet training can process previously collected trajectories while rollout collects a new batch.
\name follows asynchronous RL execution~\cite{streamrl, areal, asyncrlhf} to overlap these two workloads.
It places generation and training on shared GPUs under separate logical SM quotas, allowing both to execute concurrently.

% Role: mechanism -- describe data and weight dependencies in order.
Fig.~\ref{fig:async-pipelined-timeline} illustrates the execution order.
While $\mathit{Rollout}_i$ collects new trajectories, the trainer processes the preceding batch in $\mathit{Train}_{i-1}$.
Completed trajectories from $\mathit{Rollout}_i$ are transferred to the trainer and become available to $\mathit{Train}_i$.
The trainer processes them after its preceding update completes.
When $\mathit{Train}_{i-1}$ finishes, the runtime publishes the updated weights to generation workers for subsequent rollout work.
This ordering preserves trajectory-before-update and update-before-publication dependencies without a global rollout--training barrier.

% Role: tradeoff -- do not infer a staleness guarantee from a timeline.
Overlapping trajectory collection with policy updates introduces policy staleness.
Prior asynchronous RL systems have shown that such staleness can be managed without compromising training convergence~\cite{streamrl, areal, asyncrlhf}.
% The illustrated schedule has a one-iteration offset, but the actual policy-version lag depends on stage progress and weight publication.
% Asynchronous RL systems use staleness control and off-policy correction to manage this tradeoff~\cite{streamrl, areal, asyncrlhf}.
% The learning impact must therefore be assessed together with throughput, rather than inferred from overlap alone.
% AUTHOR CHECK: Specify the implemented weight-switch boundary, policy-version
% tracking, lag bound/backpressure rule, and off-policy correction. The current
% implementation text describes weight pushes but does not establish a lag bound.

\subsubsection{Intra-Rollout Simulation--Generation Pipeline}
\label{sec:design-intra}
% Role: opening, challenge, and design -- extend overlap from iterations to environment groups.
Within rollout, simulation and generation must also overlap to use their allocated resources concurrently.
Processing all environments as one batch forces these stages to alternate, even when rollout overlaps with training (\S\ref{sec:limitations}).
Each environment depends on its own observations and actions, but different environments can occupy different stages.
\name exploits this independence by partitioning environments into groups and pipelining simulation and generation across groups.
One group simulates its action chunk while another receives newly generated actions.

\begin{figure}[t]
\centering
\includegraphics[width=0.95\linewidth]{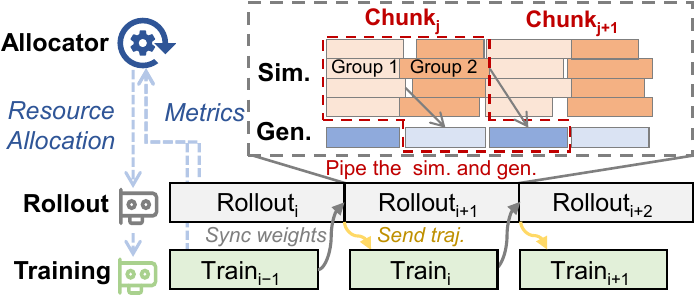}
\caption{Stage-level async-pipelined execution. Simulation and generation overlap across environment groups, while rollout and training overlap across iterations. Generation and training share GPUs under logical SM quotas; dashed arrows denote policy-weight publication.}
\label{fig:async-pipelined-timeline}
\end{figure}

% Role: mechanism -- concrete forward walk through two groups.
Consider Groups~1 and~2 in Fig.~\ref{fig:async-pipelined-timeline}.
After Group~1 executes an action chunk, the runtime places its observations in a bounded producer--consumer queue.
Generation consumes these observations and produces the next action chunk for Group~1 while Group~2 executes its available actions.
The returned actions let Group~1 resume simulation while generation serves another ready group.
Within a chunk, the step-level scheduler lets environments advance independently (\S\ref{sec:design-step}).
Simulation and generation thus overlap across groups while training consumes earlier trajectories.

% Role: advantage and limitation -- explain what pipelining can and cannot hide.
The bounded queue limits buffered observations when generation falls behind.
Group-level batching retains batched policy execution without synchronizing all rollout environments.
However, a group still needs its next actions before it can resume simulation.
Insufficient ready groups, simulator variability, or mismatched stage capacities can therefore leave pipeline bubbles.
The allocator in \S\ref{sec:design-alloc} uses the resulting stage rates to guide resource allocation.
% AUTHOR CHECK: Confirm group formation, queue capacity, and whether partial
% groups can enter generation; these parameters affect overlap and the model.

\subsubsection{Step-Level Asynchronous Orchestration}
\label{sec:design-step}
% Role: opening, challenge, and design -- derive independent steps and work-conserving rebinding.
Keeping CPU cores busy within the group pipeline requires scheduling individual environments as their work becomes ready.
Simulator-step latency varies across environments, so a barrier after every step makes fast environments wait for stragglers.
Each environment already has an action chunk and can execute its next action without waiting for other environments.
\name therefore lets environments advance independently within each chunk.
When an environment waits for new actions, its assigned cores (details in \S\ref{sec:design-alloc}) can still sit idle despite ready work elsewhere.
Dynamic core rebinding assigns these freed cores to ready workers across groups within simulation's current CPU budget.

\begin{figure}[t]
\centering
\setlength{\abovecaptionskip}{5pt}
\includegraphics[width=1\linewidth]{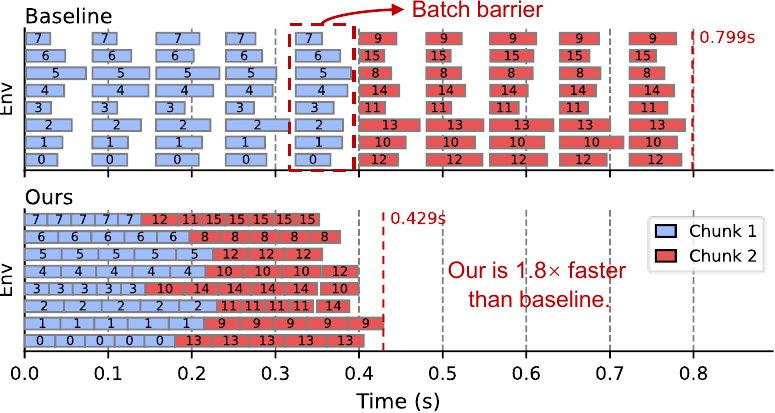}
\caption{Step-level orchestration. Environments advance independently within an action chunk, and freed CPU cores serve ready workers across groups.}
\label{fig:dynamic-step}
\vspace{-1em}
\end{figure}

% Role: mechanism -- preserve sequential actions without a cross-env barrier.
\textbf{Independent per-environment execution.}
Generation produces a sequence of actions for each environment before an action chunk begins.
An environment can therefore execute those actions in order without synchronizing with other environments at every simulator step.
As shown in Fig.~\ref{fig:dynamic-step}, each environment starts its next step as soon as its current step completes.
It becomes ready for another generation request at its action-chunk boundary, subject to the group batching in \S\ref{sec:design-intra}.
This removes within-chunk step barriers while preserving each environment's action order.

% Role: rationale -- affinity alone cannot follow ready work.
\textbf{Dynamic core rebinding.}
Independent execution can still leave an environment waiting for generation while another has actions ready to execute.
A fixed core assignment cannot redirect the waiting environment's capacity to this ready work.
To reduce CPU waiting and action-chunk tails, the scheduler follows a greedy heuristic with two principles.
(1) Assign freed cores to ready environments across groups without waiting for a group boundary.
(2) Prioritize the ready environment with the longest estimated simulator-step latency, so that likely stragglers start earlier.

% Role: definitions -- introduce the state used by the scheduling policy.
Let $s_e$ denote the number of completed steps in environment $e$'s current chunk of length $K$, and let $a_{e,s_e}$ be its next action.
The set $\mathcal{E}_{\mathrm{running}}$ tracks environments currently executing a step.
The scheduler records each environment's most recent step duration and gives priority to environments with no recorded history.
An environment is ready when it has an available action and no step in flight:
\begin{equation}
  \mathcal{E}_{\mathrm{ready}} =
  \{e \mid s_e<K,\ e\notin\mathcal{E}_{\mathrm{running}},
  \ a_{e,s_e}\text{ is available}\}.
  \label{eq:ready-env}
\end{equation}

% Role: mechanism -- follow the ready-set construction and empty-set branch.
When a core becomes free, the scheduler identifies ready environments across all groups according to Eq.~\ref{eq:ready-env}.
If no environment is ready, the core remains idle.
Otherwise, the scheduler selects the ready environment with the longest recorded step duration.
It performs this selection and marks the environment as running in a single atomic operation to prevent duplicate dispatches.
The scheduler then binds the selected environment's worker to the free core and dispatches one simulator step.

Once the step completes, the scheduler advances the environment's action index and updates its recorded step duration with the new measurement.
It then clears the environment's running status and releases the core for reassignment.
Alg.~\ref{alg:greedy-step} in Appendix~\ref{sec:appendix-core-rebinding} presents the pseudocode for single-core workers.

% Role: bridge -- concurrency alone does not determine throughput.
These mechanisms expose concurrent work while preserving each environment's action--observation dependencies.
Their realized throughput depends on the resources assigned to each stage.
We next describe the resource pool and the allocator that matches stage capacities.

\subsection{Fine-Grained Resource Management}
\label{sec:design-T1}
% Role: opening, rationale, and roadmap -- derive resource controls and allocation policy.
The asynchronous pipeline needs resource allocations that match the different demands of simulation, generation, and training.
Whole-device allocation is too coarse when one stage cannot use a full GPU, and fixed allocations cannot track changing stage costs (\S\ref{sec:challenges}).
\name therefore manages CPU cores and GPU capacity through two complementary mechanisms.
The GPU--CPU resource pool (\S\ref{sec:design-pool}) exposes sub-device capacity and the controls needed to assign it.
Elastic resource allocation (\S\ref{sec:design-alloc}) uses these controls to balance stage capacities as training progresses.
\subsubsection{GPU--CPU Resource Pool}
\label{sec:design-pool}
% Role: opening, rationale, and design -- derive the pool from stage resource requirements.
Allocating resources according to stage demand requires a representation finer than whole devices.
Generation and training share GPU execution capacity and memory, while simulation needs CPU cores and any GPU resources required by its backend.
\name represents these resources in a GPU--CPU pool with a common allocation interface.
The pool exposes GPU SM quotas, device-memory budgets, and CPU core sets while retaining their separate capacity constraints.
% The allocator uses this interface to pass stage resource budgets to the placement backends.

% Role: mechanism and tradeoff -- GPU resource representation.
\textbf{GPU pool.}
Each GPU exposes a logical resource descriptor $(s,m)$ for a stage, where $s$ is an SM quota percentage and $m$ is a device-memory budget.
Generation and training hold separate descriptors on shared GPUs; GPU-accelerated simulation also receives a descriptor where needed.
The default backend uses NVIDIA MPS to support concurrent worker processes.
% MPS quotas control execution capacity but do not provide the hardware isolation of MIG or eliminate shared-memory-bandwidth interference.
Deployments requiring hardware isolation can use pre-created MIG partitions, at the cost of coarser allocation choices.
% AUTHOR CHECK: The implementation currently sets MPS active-thread percentages
% at worker launch. Document the supported reconfiguration path before claiming
% that existing worker contexts can change quotas in place.

% Role: mechanism -- reconcile affinity with deliberate rebinding.
\textbf{CPU pool.}
The pool represents CPU capacity as cores tagged with NUMA topology.
Simulator workers are initially bound to core sets sized for their CPU requirements.
Affinity restricts uncontrolled migration during execution.
The step-level scheduler may deliberately change a worker's binding between simulator steps (\S\ref{sec:design-step}).
Thus, core rebinding changes which ready worker uses an allocated core set without changing simulation's total CPU budget.

% Role: interface -- separate budget selection from work dispatch.
\textbf{Control interface.}
The pool exposes resource descriptors and three control operations to the runtime.
\texttt{set\_quota} records a stage's target GPU SM and memory budgets for the placement backend.
\texttt{bind} assigns an initial CPU core set to a worker.
Runtime rebinding uses \texttt{sched\_setaffinity} to change the allowed core set between simulator steps.
The elastic allocator selects resource budgets, while the execution scheduler dispatches ready work within those budgets.
Backend-specific enforcement is described in \S\ref{sec:implementation}.

\subsubsection{Elastic Resource Allocation}
\label{sec:design-alloc}
% Role: opening and design -- state the goal and approach, then introduce the formulation.
To maximize end-to-end training throughput under a fixed hardware budget, we present a profile-guided elastic resource allocator.
The allocator jointly selects CPU configurations, GPU quotas, and batch sizes across stages and adjusts the allocation as stage costs change.
We first formulate the joint optimization problem, defining its throughput objective.

\begin{algorithm}[t]
\footnotesize
\caption{Elastic resource allocation: initial or updated profiles}
\label{alg:joint-opt}
\begin{algorithmic}[1]
\Require Profiles $\mathcal{P}_{\mathrm{sim}},\mathcal{P}_{\mathrm{gen}},\mathcal{P}_{\mathrm{train}}$; cores $N$; memory $M_{\mathrm{gpu}}$
\Ensure Configuration $\mathcal{C}^{*}$ with CPU/SM quotas and batch sizes
\State $c_{\mathrm{env}}^{*}\gets$ profiled per-environment core count\label{line:alloc-cores}
\State $E\gets\lfloor N/c_{\mathrm{env}}^{*}\rfloor$\label{line:alloc-concurrency}
\State $s_{\mathrm{sim}}^{*}\gets$ simulator-knee quota at $(c_{\mathrm{env}}^{*},E)$\label{line:alloc-sim-quota}
\State $T_s\gets\mathcal{P}_{\mathrm{sim}}(c_{\mathrm{env}}^{*},E,s_{\mathrm{sim}}^{*})$\label{line:alloc-sim-rate}
\State $lo\gets s_g^{\min}$; $hi\gets100-s_{\mathrm{sim}}^{*}-s_t^{\min}$\label{line:alloc-bounds}
\While{$hi-lo>1$}\label{line:alloc-loop}
  \State $s_g\gets\lfloor(lo+hi)/2\rfloor$\label{line:alloc-gen-quota}
  \State $s_t\gets100-s_{\mathrm{sim}}^{*}-s_g$\label{line:alloc-train-quota}
  \State $(B_g,B_t)\gets$ best feasible pair for $(s_g,s_t,M_{\mathrm{gpu}})$\label{line:alloc-batches}
  \State $T_r\gets\min(T_s,\mathcal{P}_{\mathrm{gen}}(s_g,B_g))$\label{line:alloc-rollout-rate}
  \State $T_t\gets\mathcal{P}_{\mathrm{train}}(s_t,B_t)$\label{line:alloc-train-rate}
  \If{$T_r>T_t$}\label{line:alloc-compare}
    \State $hi\gets s_g$\label{line:alloc-upper}
  \Else\label{line:alloc-else}
    \State $lo\gets s_g$\label{line:alloc-lower}
  \EndIf
\EndWhile
\State Evaluate both endpoints with their feasible batch-size pairs\label{line:alloc-endpoints}
\State $\mathcal{C}^{*}\gets$ endpoint configuration with higher $\Theta$\label{line:alloc-select}
\State \Return $\mathcal{C}^{*}$\label{line:alloc-return}
\end{algorithmic}
\end{algorithm}

% Role: objective -- define comparable rates and the feasible configuration.
\textbf{Throughput objective.}
Let $\mathcal{C}=(\mathcal{C}_{\mathrm{sim}},\mathcal{C}_{\mathrm{gen}},\mathcal{C}_{\mathrm{train}})$ denote a complete configuration.
These stage configurations specify CPU-core counts, environment concurrency, GPU SM quotas, and batch sizes where applicable.
We express all stage rates in equivalent rollout chunk-steps per second.
Simulation rates account for the actions in each chunk; training rates account for the prescribed updates and sample reuse.
Under steady-state overlap, the predicted end-to-end throughput is
\begin{equation}
\begin{aligned}
T_r(\mathcal{C}) &=
\min\{T_{\mathrm{sim}}(\mathcal{C}_{\mathrm{sim}}),
       T_{\mathrm{gen}}(\mathcal{C}_{\mathrm{gen}})\},\\
\Theta(\mathcal{C}) &=
\min\{T_r(\mathcal{C}),T_{\mathrm{train}}(\mathcal{C}_{\mathrm{train}})\}.
\end{aligned}
\label{eq:theta}
\end{equation}
The simulation--generation pipeline limits the rollout rate $T_r$ to the slower of its two stages (\S\ref{sec:design-intra}).
Since rollout further overlaps with training (\S\ref{sec:design-inter}), the end-to-end rate is limited by the slower of rollout and training, yielding the second minimum.
% This model assumes enough in-flight work to sustain overlap.
% Queueing and GPU interference can reduce achieved throughput.
% Role: constraints -- connect the model to the resource pool.
With this objective, the allocator seeks $\mathcal{C}^{*}\in\arg\max_{\mathcal{C}\in\mathcal{F}}\Theta(\mathcal{C})$, where $\mathcal{F}$ contains configurations that satisfy the GPU and CPU resource constraints.
% On a shared GPU, the SM quotas satisfy $s_{\mathrm{sim}}+s_g+s_t\leq100$.
% On each shared GPU, the combined memory demand of resident models, runtime buffers, and batch-dependent state must fit the device budget $M_{\mathrm{gpu}}$.
% In addition, the simulator must receive a nonzero GPU quota whenever its backend requires GPU resources.
% On the CPU side, worker bindings must fit the simulation core budget $N$.
% Together, these constraints enforce resource feasibility on the devices hosting the corresponding workers.
% AUTHOR CHECK: Verify that measured training rates are normalized for actual
% sample reuse and that memory profiles include simultaneous residency and
% weight-transfer buffers.

% Role: mechanism -- obtain performance inputs without end-to-end enumeration.
\textbf{Stage performance characterization.}
Direct end-to-end search is expensive because quotas, environment concurrency, and batch sizes form a coupled configuration space.
\name offline profiles each stage and stores its throughput in lookup tables $\mathcal{P}_{\mathrm{sim}}$, $\mathcal{P}_{\mathrm{gen}}$, and $\mathcal{P}_{\mathrm{train}}$.
Simulator profiling varies cores per environment, concurrent environments, and any required SM quota.
Generation profiling replays observation batches under different SM quotas and batch sizes.
Training profiling replays rollout batches using FSDP under the corresponding quotas and training batch sizes.
The isolated profiles provide initial capacity estimates; runtime measurements reveal deviations during shared execution.

% Role: mechanism and rationale -- explain the allocation search in order.
\textbf{Initial allocation.}
To increase end-to-end throughput without enumerating all configurations, the allocator follows two principles.
(1) Establish simulator capacity using a profiled CPU configuration and the smallest GPU quota that reaches its throughput knee.
(2) Shift the remaining GPU capacity toward the slower of rollout and training while choosing batch sizes within the memory budget.

Alg.~\ref{alg:joint-opt} gives the full procedure.
The allocator first uses the profiles to choose the cores per environment, $c_{\mathrm{env}}^{*}$ (line~\#\ref{line:alloc-cores}).
Given the core budget $N$, it can run $E=\lfloor N/c_{\mathrm{env}}^{*}\rfloor$ environments concurrently (line~\#\ref{line:alloc-concurrency}).
It reserves the smallest simulator SM quota beyond which additional SMs bring little throughput gain, and looks up the resulting simulator capacity $T_s$ (lines~\#\ref{line:alloc-sim-quota}--\ref{line:alloc-sim-rate}).
The remaining SMs are shared by generation and training, with each stage receiving at least its minimum quota, $s_g^{\min}$ and $s_t^{\min}$ (line~\#\ref{line:alloc-bounds}).
% AUTHOR CHECK: Distinguish simultaneous CPU execution slots E from the total
% number of environments across pipeline groups in the implementation.

% Role: mechanism -- explain candidate evaluation and each bisection branch.

The allocator uses binary search to balance rollout and training throughput.
At each step, it tests the midpoint of the generation-quota range, assigns the remaining SMs to training, and chooses the batch-size pair that gives the highest predicted end-to-end throughput within the memory budget $M_{\mathrm{gpu}}$ (lines~\#\ref{line:alloc-gen-quota}--\ref{line:alloc-batches}).
It estimates rollout throughput as the lower of simulation and generation throughput, then compares it with training throughput (lines~\#\ref{line:alloc-rollout-rate}--\ref{line:alloc-compare}).
If rollout is faster, the search shifts toward smaller generation quotas to give training more SMs (lines~\#\ref{line:alloc-compare}--\ref{line:alloc-upper}); otherwise, it shifts toward larger generation quotas (lines~\#\ref{line:alloc-else}--\ref{line:alloc-lower}).
When the range is at most one percentage point wide (line~\#\ref{line:alloc-loop}), the allocator checks both endpoints and returns the configuration with higher $\Theta$ (lines~\#\ref{line:alloc-endpoints}--\ref{line:alloc-return}).

% Role: assumption -- state the search condition after the forward walkthrough.
% The bisection assumes that increasing $s_g$ does not reduce rollout capacity and does not increase training capacity.
% This trend must hold over the feasible interval; shared-memory constraints and interference can invalidate it.
% The simulator-first decomposition reduces search cost without establishing a global optimum over all CPU--GPU configurations.
% AUTHOR CHECK: Validate the feasible interval and monotonicity used by bisection.
% Specify handling of infeasible splits or non-monotonic measured profiles;
% do not claim uniqueness or global optimality without this evidence.

% Role: mechanism -- reuse the allocator as workload demand changes.
\textbf{Runtime adaptation.}
Policy evolution and simulator variability can shift the balance between rollout and training.
\name monitors a 20-step moving average of observed rollout and training throughputs.
It uses runtime measurements to refine the lookup tables underlying the current allocation.
When the gap exceeds 15\% of the faster stage for 20 consecutive steps, the runtime reruns Algorithm~\ref{alg:joint-opt} with updated lookup tables.
The resulting configuration is passed to the resource pool for application.
Initial placement and subsequent adjustment thus use the same allocation procedure.

% Role: tradeoff -- separate budget adaptation from fast work scheduling.
% The sustained-imbalance trigger filters transient fluctuations before invoking another allocation search.
% This adaptation changes stage budgets, whereas core rebinding (\S\ref{sec:design-step}) schedules environments within simulation's current CPU budget.
% Their costs differ: rebinding occurs between simulator steps, while a new GPU allocation requires backend-specific reconfiguration.
% The runtime must preserve in-flight work when applying such changes.
% AUTHOR CHECK: Document the monitoring step unit, simulator-profile update,
% GPU reconfiguration boundary, and measured transition overhead. Launch-time
% MPS environment variables alone do not explain online changes to live workers.

\subsection{Runtime and Programming Support}
\label{sec:design-prog}
% Role: opening, rationale, and interface -- connect design mechanisms to user configuration.
\name provides a Python programming model for resource configuration and training orchestration (Listing~\ref{lst:prog-snippet}).
Users configure the policy, simulator, and resource pool, then express a training loop that feeds execution statistics back to the allocator.
The runtime implements these operations through resource profiling, elastic placement, and asynchronous worker scheduling.
It reuses RLinf's WorkerGroup dispatch, communication channels, and cluster launch to execute and connect workers.

{\lstset{aboveskip=2pt,belowskip=2pt,abovecaptionskip=2pt,belowcaptionskip=1pt}
\begin{lstlisting}[style=embodirl,language=Python,float=t,floatplacement=t,
    basicstyle=\ttfamily\footnotesize,
    numbersep=3pt,xleftmargin=10pt,
    label=lst:prog-snippet,
    caption={\name programming model for resource configuration and training orchestration. The runtime manages elastic allocation, asynchronous execution, and step-level scheduling.}]
from ebrl import ResourcePool, EmbodiPlacement, profile
class EmbodiRLRunner:
    def __init__(self):
        self.pool = ResourcePool(num_gpus=4, gpu_pool="mps", cpu_pool="numa")
        self.prof = profile(self.pool, model="pi05", env="libero",
            sm_grid=[20, 40, 60, 80], batch_grid=[8, 16, 32, 64])
        self.placement = EmbodiPlacement(pool=self.pool, profile=self.prof)
        self.placement.apply()
    def run(self, num_iters):
        for it in range(num_iters):
            stats = train_one_iter()
            self.placement.balance(stats).apply()
\end{lstlisting}}

% Role: mechanism -- initialization creates the budgets consumed by execution.
\textbf{Initialization.}
The user creates a \texttt{ResourcePool} to expose available CPU and GPU capacity, and invokes \texttt{profile} with the selected policy and simulator to collect stage performance tables.
\texttt{EmbodiPlacement} computes the initial allocation (\S\ref{sec:design-alloc}).
Its \texttt{apply} method passes the resulting CPU bindings and GPU quotas to the placement backend for execution-worker launch.

% Role: mechanism -- distinguish runtime scheduling from budget changes.
\textbf{Training and runtime coordination.}
The training loop passes throughput statistics to \texttt{balance}, which triggers elastic allocation when a sustained imbalance is detected.
The subsequent \texttt{apply} call passes changed configurations to the placement backend.
The loop expresses training progress without imposing a barrier between rollout and training.
Workers exchange observations, actions, trajectories, and weights through the asynchronous pipelines in \S\ref{sec:design-T2}.
The step-level scheduler dispatches ready simulator work within the current CPU allocation.
\S\ref{sec:implementation} describes the integration with RLinf, PyTorch FSDP, and the resource-control backends.

\vspace{-0.5em}
\section{Implementation}
\label{sec:implementation}
% \vspace{-0.5em}

We implement \name as a fine-grained resource and runtime orchestration layer on top of RLinf~\cite{rlinf,rlinfvla}, the SOTA embodied RL framework.
The codebase primarily uses Python, totaling $\sim$6.0K LOC, and reuses RLinf's WorkerGroup dispatch, Ray launch, data channels, and training path, changing only resource placement and rollout process.

The GPU--CPU resource pool (\S\ref{sec:design-pool}) is a scheduler-side placement backend that resolves the user's config into one \texttt{WorkerResourceBinding} per worker rank.
GPU SM partitioning is enforced through MPS active-thread percentages, set per process via the \texttt{CUDA\_MPS\_\allowbreak ACTIVE\_\allowbreak THREAD\_\allowbreak PERCENTAGE} environment variable injected at worker launch, and falls back to pre-created MIG UUIDs through \texttt{CUDA\_\allowbreak VISIBLE\_\allowbreak DEVICES} when hardware isolation is required; CPU pinning is enforced through Linux \texttt{os.\allowbreak sched\_\allowbreak setaffinity\allowbreak (pid,\allowbreak\ cores)} on each worker and its simulator subprocesses.
\name does not manage the MPS daemon or mutate MIG instances at runtime, so deployments only need a one-time MPS/MIG setup.
The joint allocator (\S\ref{sec:design-alloc}) is an offline-profile solver invoked at warm-up that sweeps simulator, generation, and FSDP-sharded training throughput under varying CPU/SM quotas and runs Algorithm~\ref{alg:joint-opt} to find the binding plan.
The runtime rollout scheduler (\S\ref{sec:design-T2}) implements two mechanisms.
Stage-level async-pipelined execution (\S\ref{sec:design-async}) runs simulation workers continuously via a bounded producer--consumer queue between env workers and generation, with generation and training co-located under MPS and synchronized via CUDA IPC events for non-blocking weight pushes.
Step-level asynchronous orchestration (\S\ref{sec:design-step}) inherits per-env optimal core counts from \S\ref{sec:design-alloc}; at runtime, environments advance asynchronously within an action chunk, and the scheduler rebinds each freed core to the slowest ready env via \texttt{os.\allowbreak sched\_\allowbreak setaffinity} (<1ms overhead).

\begin{figure*}[t]
\centering
\setlength{\abovecaptionskip}{5pt}
\includegraphics[width=\linewidth]{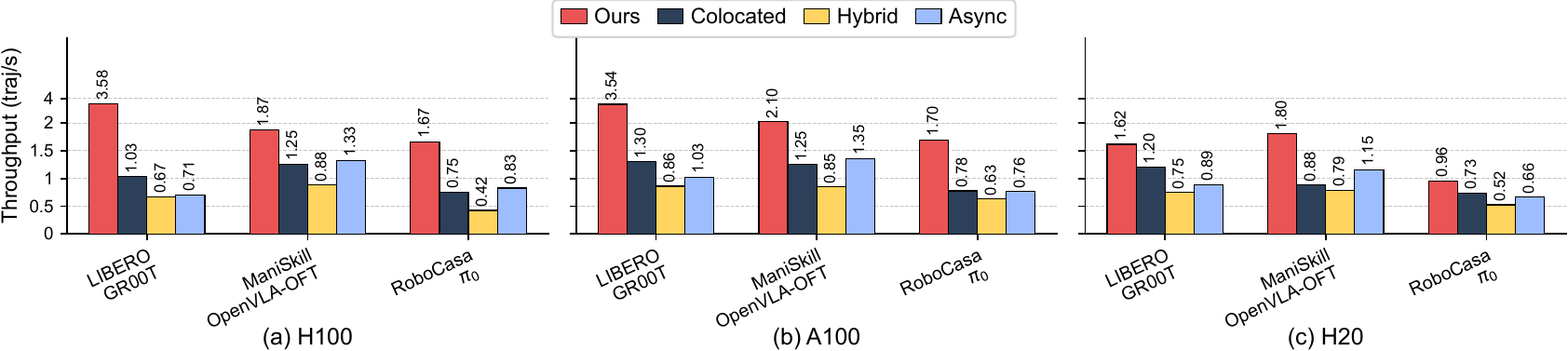}
\caption{End-to-end trajectory throughput for LIBERO/GR00T, ManiSkill/OpenVLA-OFT, and RoboCasa/OpenPI on H100, A100, and H20.}
\label{fig:e2e_throughput}
\end{figure*}

\section{Evaluation}\label{sec:evalutaion}\label{sec:eval}
\name improves embodied RL throughput by combining fine-grained resource allocation with asynchronous pipelining.
Our evaluation covers five simulator-policy configurations across H100, A100, and H20 testbeds, with an additional RTX 4090 pool for BEHAVIOR.
The key findings are as follows.

\noindent\scalebox{0.8}{$\bullet$} \name achieves 1.30$\times$--3.47$\times$ the throughput of the SOTA RL systems in each configuration and retains its advantage with GRPO. (\S\ref{sec:E2EPerf})

\noindent\scalebox{0.8}{$\bullet$} Removing resource allocation, group pipelining, or asynchronous execution reduces throughput by approximately 10\%, 31\%, and 26\%, respectively. 
% CPU affinity improves simulator throughput by 1.38$\times$--1.97$\times$ across the three benchmarks. (\S\ref{sec:ComponentwiseAnalysis})

\noindent\scalebox{0.8}{$\bullet$} \name reaches approximately 98\% success in the convergence experiment and maintains 1.59--1.62 trajectories/s .(\S\ref{sec:E2EPerf})

\subsection{Experimental Setup}\label{sec:ExperimentalSetup}
\textbf{Policies.} We evaluate \name on four representative embodied policies covering the VLA and WAM families.
(1) \textit{OpenPI}~\cite{pi0} and (2) \textit{GR00T}~\cite{groot} pair a VLM backbone with a diffusion action expert.
(3) \textit{OpenVLA-OFT}~\cite{openvla} is a VLA optimized for efficient fine-tuning.
(4) \textit{DreamZero}~\cite{dreamzero} is a Diffusion-Transformer WAM that jointly denoises future frames and actions.
Unless noted, the VLM backbone is updated with LoRA adapters while the action head/expert is fully fine-tuned, following each policy's recipe.

\textbf{Simulators and configurations.} We use CPU-based and GPU-accelerated simulators to cover different resource demands (Table~\ref{tab:sim_platforms}).
\textit{LIBERO}~\cite{libero} and \textit{RoboCasa}~\cite{robocasa} rely mainly on CPU-side simulation, while \textit{ManiSkill}~\cite{maniskill} and \textit{BEHAVIOR}~\cite{behavior} use GPU-accelerated simulation.
The five configurations are LIBERO-Object/GR00T, ManiSkill/OpenVLA-OFT, RoboCasa/OpenPI, BEHAVIOR/OpenPI, and LIBERO/DreamZero.
The main comparison uses PPO~\cite{ppo}.
We additionally evaluate GRPO on RoboCasa and ManiSkill.

\textbf{Testbeds.} The main comparison uses eight-GPU H100 with Intel(R) Xeon(R) Platinum 8468 CPU, A100 with Intel Xeon Gold 6348 CPU, and H20 with Intel(R) Xeon(R) Platinum 8480+ CPU testbeds.
The GR00T, OpenVLA-OFT, and OpenPI configurations run on all three platforms; LIBERO with DreamZero is additionally evaluated on A100.
For BEHAVIOR, which requires RT cores for rendering, we use RTX 4090 and A100 GPUs on two machines connected over a switched network with measured bandwidth above 30\,Gbps.
The scaling experiment extends to 16 GPUs.
% AUTHOR CHECK: The main figure identifies H100/A100/H20, whereas the old
% setup listed A100/A800/H20. Restore CPU models and GPU memory capacities
% after verifying their mapping against the run configurations.

\textbf{Baselines and metrics.} We compare against two execution modes of RLinf~\cite{rlinf,rlinfvla} and an asynchronous system~\cite{rlvla3}.
\textit{Colocated} time-shares every GPU across simulation, generation, and training.
\textit{Hybrid} pipelines simulation and generation within rollout but retains a synchronous barrier before training.
\textit{Async} disaggregates rollout and training onto separate GPU pools and overlaps them with off-policy correction.
All baselines use RLinf's recommended settings.
We report end-to-end throughput in completed trajectories per second, following the throughput bar charts, and simulator throughput in steps/s for the CPU-affinity microbenchmark.
Convergence is measured by evaluation success rate versus wall-clock training time.
% Ratios use the underlying values in ../eval_fig plotting scripts, before
% rounding the bar labels. No chunk-step conversion is used.

\subsection{End-to-End Performance}\label{sec:E2EPerf}
\name achieves higher throughput across the evaluated workloads and hardware platforms while retaining comparable final success in the convergence experiment.
Unless specified otherwise, the end-to-end experiments use PPO.

\begin{figure}[t]
\centering
\setlength{\abovecaptionskip}{3pt}
\begin{minipage}[t]{0.48\linewidth}
\centering
\vspace{0pt}
\includegraphics[width=\linewidth]{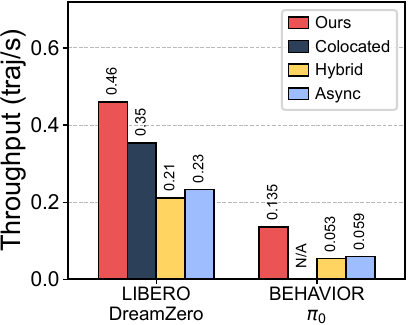}
\caption{Throughput of LIBERO/DreamZero (A100) and BEHAVIOR/OpenPI (RTX 4090--A100).}
\label{fig:e2e_throughput_dreamzero_behavior}
\end{minipage}
\hfill
\begin{minipage}[t]{0.48\linewidth}
\centering
\vspace{0pt}
\includegraphics[width=\linewidth]{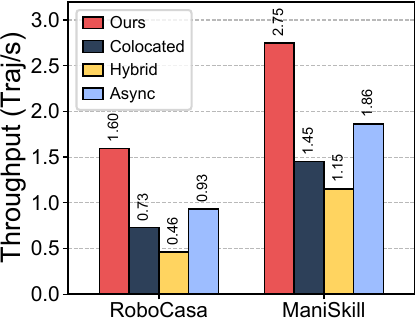}
\caption{GRPO throughput on RoboCasa and ManiSkill.}
\label{fig:grpo_throughput}
\end{minipage}
\end{figure}

\noindent\textbf{Throughput across configurations and testbeds.}
\name outperforms all three baselines across the representative VLA configurations in Fig.~\ref{fig:e2e_throughput}.
We evaluate GR00T with LIBERO, OpenVLA-OFT with ManiSkill, and OpenPI with RoboCasa on H100, A100, and H20, yielding nine workload--testbed combinations.
Across these combinations, \name achieves 1.31$\times$--3.47$\times$ the throughput of the strongest baseline.
Its mean speedups over Colocated, Hybrid, and Async are 2.05$\times$, 3.00$\times$, and 2.28$\times$.
On H100, the gains over the strongest baseline are 3.47$\times$, 1.41$\times$, and 2.02$\times$ for LIBERO, ManiSkill, and RoboCasa, respectively.
On A100, the gains range from 1.55$\times$ to 2.72$\times$; on H20, they range from 1.31$\times$ to 1.56$\times$.
We further validate \name with the DreamZero WAM on LIBERO and with OpenPI on BEHAVIOR, which requires RT-capable GPUs for rendering (Fig.~\ref{fig:e2e_throughput_dreamzero_behavior}).
On LIBERO/DreamZero with A100 GPUs, \name achieves 1.30$\times$, 2.18$\times$, and 1.97$\times$ the throughput of Colocated, Hybrid, and Async, respectively.
For BEHAVIOR on the mixed RTX 4090--A100 testbed, the speedups over Hybrid and Async are 2.55$\times$ and 2.29$\times$.\footnote{Colocated is unavailable because A100 GPUs lack the RT cores needed for BEHAVIOR rendering, preventing all three stages from running on each GPU.}
These results extend the throughput advantage to the WAM policy family and to BEHAVIOR's heterogeneous rendering and learning resources.

\begin{figure}[t]
\centering
\setlength{\abovecaptionskip}{3pt}
\includegraphics[width=0.8\linewidth]{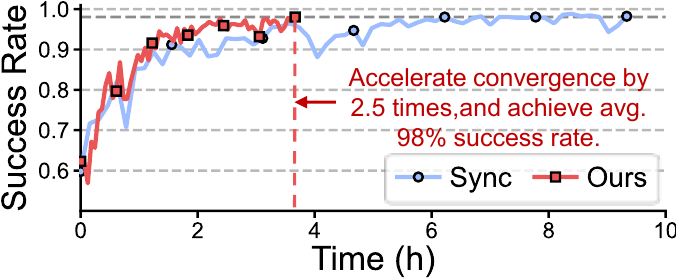}
\caption{Evaluation success rate versus wall-clock training time for Sync and \name on LIBERO-Object/GR00T with PPO.}
\vspace{-2em}
\label{fig:model_convergence}
\end{figure}

These gains reflect the complementary roles of resource sharing and overlap.
Colocated serializes stages with different CPU and GPU demands, leaving resources idle when the active stage cannot use them.
Hybrid overlaps work within rollout but still waits at the rollout--training boundary.
Async removes that boundary but dedicates separate GPU pools to rollout and training, limiting capacity sharing between the stages.
\name combines overlap with fine-grained allocation, allowing concurrent stages to use the same resource pool and reducing the idle capacity caused by these execution constraints.

\begin{figure*}[t]
\centering
\setlength{\abovecaptionskip}{3pt}
% Equal-height content boxes align the starts of all five captions.
\begin{minipage}[t]{0.19\linewidth}
\centering
\vspace{0pt}
\begin{minipage}[t][0.9\linewidth][t]{\linewidth}
\centering
\vspace{0pt}
\includegraphics[width=\linewidth]{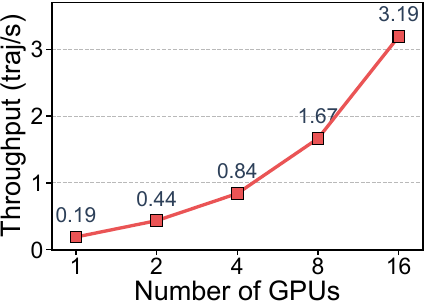}
\end{minipage}\par
\caption{Throughput of \name with 1--16 GPUs.}
\label{fig:resource_scaling}
\end{minipage}
\hfill
\begin{minipage}[t]{0.19\linewidth}
\centering
\vspace{0pt}
\begin{minipage}[t][0.9\linewidth][t]{\linewidth}
\centering
\vspace{0pt}
\includegraphics[width=\linewidth]{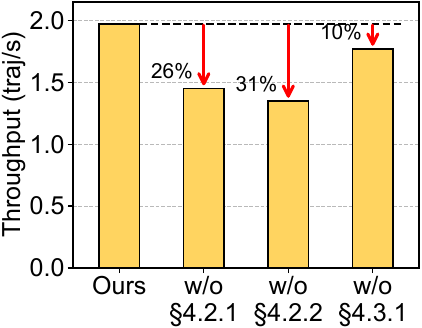}
\end{minipage}\par
\caption{Throughput contribution of the three mechanisms.}
\label{fig:eval_ablation}
\end{minipage}
\hfill
\begin{minipage}[t]{0.19\linewidth}
\centering
\vspace{0pt}
\begin{minipage}[t][0.9\linewidth][t]{\linewidth}
\centering
\vspace{0pt}
\includegraphics[width=\linewidth]{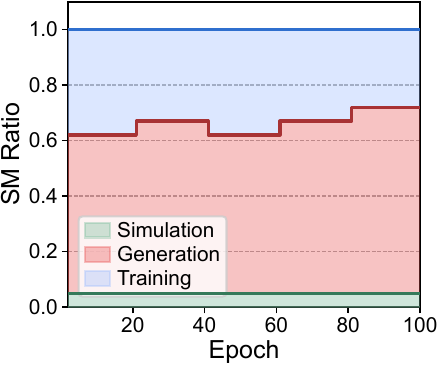}
\end{minipage}\par
\caption{SM shares derived from training times.}
\label{fig:allocation_timeline}
\end{minipage}
\hfill
\begin{minipage}[t]{0.19\linewidth}
\centering
\vspace{0pt}
\begin{minipage}[t][0.9\linewidth][t]{\linewidth}
\centering
\vspace{0pt}
\includegraphics[width=\linewidth]{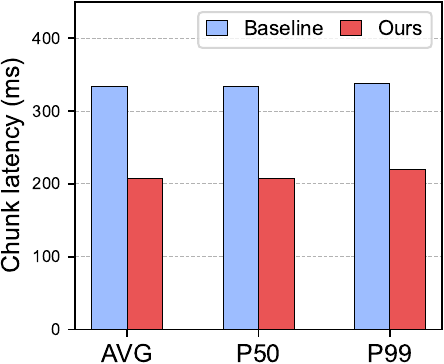}
\end{minipage}\par
\caption{Average, P50, and P99 action-chunk latency.}
\label{fig:chunk_latency}
\end{minipage}
\hfill
\begin{minipage}[t]{0.19\linewidth}
\centering
\vspace{0pt}
\begin{minipage}[t][0.9\linewidth][t]{\linewidth}
\centering
\vspace{0pt}
\small
\setlength{\tabcolsep}{3pt}
\begin{tabular}{@{}cc@{}}
\toprule
Staleness & \shortstack{Throughput\\(traj/s)} \\
\midrule
1 & 1.62 \\
2 & 1.60 \\
4 & 1.59 \\
8 & 1.60 \\
\bottomrule
\end{tabular}
\end{minipage}\par
\caption{Throughput sensitivity to policy staleness.}
\label{fig:staleness_throughput}
\end{minipage}
\end{figure*}

\noindent\textbf{Model convergence.}
On LIBERO-Object/GR00T with PPO, \name completes 60 epochs approximately 2.5$\times$ faster than Sync, with both reaching approximately 98\% final success (Fig.~\ref{fig:model_convergence}).
The throughput advantage thus translates into shorter training time without compromising final task success in this experiment.
This observation is consistent with prior asynchronous RL systems~\cite{streamrl,areal,asyncrlhf}: overlapping rollout and training can improve training efficiency while preserving effective policy learning.

\noindent\textbf{Throughput with GRPO.}
\name retains its throughput advantage when the RL algorithm changes to GRPO.
We compare the four systems on RoboCasa/OpenPI (CPU-based simulator) and ManiSkill/OpenVLA-OFT (GPU-based simulator) under GRPO in Fig.~\ref{fig:grpo_throughput}.
On RoboCasa, \name achieves 2.20$\times$, 3.45$\times$, and 1.71$\times$ the throughput of Colocated, Hybrid, and Async, respectively; on ManiSkill, the corresponding speedups are 1.90$\times$, 2.39$\times$, and 1.48$\times$.
The allocator uses stage performance profiles to account for the algorithm's rollout and training demands, while asynchronous pipelining overlaps the resulting workloads.

% AUTHOR CHECK: Identify the Sync mode and document the sample budget per
% epoch, off-policy correction, and observed policy-version lag.

\noindent\textbf{Resource scaling.}
\name sustains throughput growth as the GPU count increases.
Fig.~\ref{fig:resource_scaling} evaluates 1, 2, 4, 8, and 16 GPUs.
The plotted throughputs are 0.19, 0.44, 0.84, 1.67, and 3.19, respectively.
\name can well handle the increased capacity because its allocator balances stage capacities and its asynchronous pipelining exposes concurrent work.
Sharing resource in the same GPU avoids cross-machine communication, further ensuring scalability.
% Using the unrounded plotting data, the 16-GPU result is 16.65$\times$ the single-GPU result and 7.31$\times$ the two-GPU result.
% The latter corresponds to 91.4\% scaling efficiency from 2 to 16 GPUs.
% This trend is consistent with allocation that balances stage capacities and pipelining that exposes concurrent work as more resources become available.
% AUTHOR CHECK: The scaling plot/source specify GPU counts and throughput
% values, but not the unit, workload, hardware model, or batch policy.
% Confirm these before adding them; the plot contains no baseline curves.

\subsection{Comprehensive Component-wise Analysis}\label{sec:ComponentwiseAnalysis}
Resource allocation and asynchronous pipelining make complementary contributions to \name's throughput.
Unless specified otherwise, component experiments use RoboCasa/OpenPI on the H100 testbed.

\noindent\textbf{Ablation.}
Each component in \name contributes to system performance.
Individually disabling elastic resource allocation, intra-rollout simulation--generation pipelining, and inter-iteration rollout--training overlap reduces throughput by approximately 10\%, 31\%, and 26\%, respectively (Fig.~\ref{fig:eval_ablation}).
Elastic resource allocation (\S\ref{sec:design-alloc}) matches CPU and GPU resources to stage demands, reducing the capacity imbalance of a fixed split.
Intra-rollout pipelining (\S\ref{sec:design-intra}) runs simulation and generation concurrently across environment groups, avoiding idle time from alternating stages.
Inter-iteration overlap (\S\ref{sec:design-inter}) trains on previously collected trajectories while the next rollout proceeds, removing waits at the rollout--training barrier.
% AUTHOR CHECK: Record the fixed allocation and exact ablation configuration;
% the full-system value here differs from the main throughput experiment.

\noindent\textbf{Resource allocation.}\label{sec:ExperimentUtilization}
\name adjusts GPU shares as stage loads change during learning, using measured training times (Fig.~\ref{fig:allocation_timeline}).
As the policy improves and training becomes less costly, the allocator shifts capacity from training to generation.
When policy fluctuations increase training costs, runtime measurements trigger reallocation back to training.
These adjustments keep resource shares aligned with stage demands and help balance rollout and training capacities.

% \noindent\textbf{Resource utilization.}
% \name targets unused capacity exposed by heterogeneous stage demands through fine-grained sharing.
% Fig.~\ref{fig:utilization} characterizes generation and training through SM-occupancy CDFs and CPU activity through counts of cores above 50\% and 80\% utilization.
% The median SM occupancy is approximately 6\% for generation and 53\% for training.
% For much of the CPU trace, only approximately 20--30 of the 112 cores exceed 50\% utilization, and fewer exceed 80\%.
% These profiles reveal capacity that coarse stage placement can leave unused, motivating GPU sharing and reassignment of idle CPU cores to ready simulation work.
% They characterize resource demand rather than a four-system comparison of aggregate CPU and GPU utilization.

% \begin{figure}[t]
% \centering
% \setlength{\abovecaptionskip}{3pt}
% \includegraphics[width=0.4\linewidth]{figs/sm_cpu_utilization.pdf}
% \caption{Stage SM-occupancy distributions and counts of CPU cores exceeding two utilization thresholds.}
% \label{fig:utilization}
% \end{figure}

\noindent\textbf{Core rebinding.}
Core rebinding reduces action-chunk latency while maintaining stable execution.
Compared with the batch-step baseline, \name achieves 1.61$\times$ speedups in average and P50 chunk latency and a 1.53$\times$ speedup at P99 (Fig.~\ref{fig:chunk_latency}).
% The baseline's closely clustered latencies reflect its batch-step synchronization: environments advance together and wait for the slowest step in each batch, aligning their completion times but prolonging chunk execution.
\name reduces this waiting by allowing environments to progress independently and rebinding freed CPU cores to ready workers.
P99 remains only 6.3\% above P50, indicating limited tail variability.

\noindent\textbf{Sensitivity to policy staleness.}\label{sec:ExperimentStaleness}
\name maintains similar throughput across the tested policy staleness settings.
\footnote{Staleness is the number of policy versions by which the model used to collect a trajectory lags behind the trainer's model when that trajectory is used for training.}
We vary staleness over 1, 2, 4, and 8 on RoboCasa/OpenPI while keeping the resource allocation and learning workload fixed.
Throughput varies by less than 2\% relative to staleness 1 (Fig.~\ref{fig:staleness_throughput}).
% The full range is only 0.03 trajectories/s, or 1.9\% of the throughput at staleness 1.
\name balances rollout and training capacities.
% This limited variation is consistent with balanced rollout and training capacities.
Once their execution overlaps, allowing more policy lag provides little additional throughput.
% The experiment measures throughput sensitivity, while Fig.~\ref{fig:model_convergence} separately evaluates learning behavior for the reported convergence run.

\begin{figure}[t]
\centering
\setlength{\abovecaptionskip}{3pt}
\includegraphics[width=\linewidth]{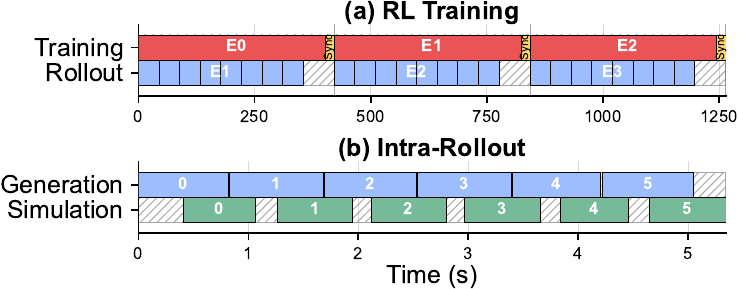}
\caption{Gantt charts of (a) rollout--training overlap across epochs and (b) generation--simulation pipelining within a rollout epoch. Hatched regions indicate idle time.}
\vspace{-1em}
\label{fig:gantt}
\end{figure}

\noindent\textbf{Async-pipelined execution.}
Fig.~\ref{fig:gantt}(a) shows training on previously collected trajectories overlapping rollout across three epochs (E1--E3).
Rollout finishes first and waits for the policy update and weight synchronization before the next epoch.
Fig.~\ref{fig:gantt}(b) shows generation and simulation overlapping across environment groups over six action chunks (0--5).
Idle time averages 7\% and 15\% across the two lanes in (a) and (b), respectively, over the illustrated windows.

% \begin{table}[h]
% \centering
% \small
% \caption{Simulator throughput with our optimal per-environment CPU bindings and default OS scheduling.}
% \label{tab:cpu_affinity_comparison}
% \begin{tabular}{@{}lcrr@{}}
% \toprule
% Benchmark & \shortstack{Cores/env} & \multicolumn{2}{c}{Throughput (steps/s)} \\
% \cmidrule(l){3-4}
%  & & Ours & OS default \\
% \midrule
% LIBERO & 1 & \textbf{1155.9} & 838.2 \\
% RoboCasa & 1 & \textbf{700.90} & 355.9 \\
% BEHAVIOR & 3 & \textbf{670.85} & 420.3 \\
% \bottomrule
% \end{tabular}
% \end{table}

% \noindent\textbf{Simulation performance.}
% Explicit CPU affinity improves simulator throughput and reduces latency.
% Compared with default OS scheduling, our CPU bindings improve simulator throughput by factors of 1.38$\times$, 1.97$\times$, and 1.60$\times$ on LIBERO, RoboCasa, and BEHAVIOR, respectively (Table~\ref{tab:cpu_affinity_comparison}).
% These benefits are due to CPU binding reducing scheduling preemption between environments.
% \noindent
% \textbf{Benefit of reset optimization.}
% % Our implementation further accelerates LIBERO resets by avoiding redundant hard seeding when resetting the same scene.
% To isolate the benefit of reset optimization, we compare \name with and without it on LIBERO-Object/GR00T, keeping all other settings unchanged.
% Fig.~\ref{fig:reset_optimization} shows that disabling the optimization reduces \name's throughput by \tododata{XX}\% to \tododata{X.XX} trajectories/s, while still achieving $1.2\times$ the throughput of the strongest baseline.
% The optimization avoids redundant hard seeding for the same scene, reducing reset latency and leaving more CPU time for simulation computation.

% AUTHOR CHECK: Document version tagging and lag control. This throughput
% sweep does not establish equal convergence at every staleness setting.

\section{Related Work}\label{sec:RelatedWorks}
% \vspace{-0.5em}
\textbf{Embodied models and simulators} improve robot capabilities and the efficiency of interaction data collection.
Vision-language-action models~\cite{rt2, openvla, pi0, groot, hpt, vlasurvey} connect pretrained vision-language representations to action generation.
World action models~\cite{dreamzero} further model future states alongside actions, while diffusion transformer policies~\cite{rdt1b} support bimanual manipulation.
RL post-training~\cite{vla2rl, dppo} further adapts these policies through closed-loop interaction with simulators~\cite{mujoco, robosuite, physx, sapien, isaacgym, isaaclab}; GPU-parallelized simulation~\cite{Maniskill3} accelerates this data collection.
Beyond model and simulator improvements, policy optimization and adaptive rollout methods~\cite{least, srpo, bhatia2022adaptive} improve the effectiveness of collected interactions.
% \name complements these efforts by coordinating the CPU and GPU resources used for simulation, generation, and training.

\textbf{Embodied RL systems} improve training efficiency through workflow placement and rollout scheduling.
RLinf~\cite{rlinf, rlinfvla, rlinf_user} supports colocated, disaggregated, and hybrid execution, including hybrid allocation for GPU-parallelized simulators.
Other systems~\cite{dvla, rlvla3, simplevlarl, unilab} explore asynchronous execution, pipelining and workload rebalancing to alleviate the long-tail problem and enhance training throughput.
Beyond stage placement and workload scheduling, \name achieves a fine-grained asynchronous execution mode with resource management.

% It further advances environments independently within action chunks and rebinds freed CPU cores to ready workers across groups.

% \textbf{Asynchronous LLM RL} reduces waiting between rollout and policy updates.
% HybridFlow~\cite{verl} supports distributed RLHF workflows, while VerlTool~\cite{verltool} extends RL training to agents that interact with tools.
% StreamRL~\cite{streamrl}, AReaL~\cite{areal}, and asynchronous RLHF~\cite{asyncrlhf} overlap rollout and training while managing the resulting policy staleness.
% \name adopts this inter-iteration overlap and further addresses the simulation--generation dependencies within embodied rollout.
% It pipelines the two stages across environment groups and allocates shared CPU and GPU capacity according to their different processing rates.

\textbf{Fine-grained resource management} improves utilization through resource allocation and hardware sharing.
Synergy~\cite{Synergy} allocates CPU cores and memory according to job sensitivity, while GoldMiner~\cite{Goldminer} elastically scales CPU workers for data preprocessing.
For GPU sharing, MPS and MIG~\cite{cuda_mps, nvidia_mig} provide spatial multiplexing mechanisms, GMI-DRL~\cite{gmidrl} applies GPU multiplexing to DRL, and Orion~\cite{Orion} schedules kernels to manage interference.
% Embodied RL additionally requires resource allocation to account for the coupled rates of simulation, generation, and training.
\name uses stage throughput profiles to jointly select CPU configurations, GPU quotas, and batch sizes, then adjusts allocations as stage costs change.

% \vspace{-1em}
\section{Conclusion}
% \vspace{-0.5em}
In this paper, we study resource underutilization in embodied RL training.
We find that existing systems allocate GPUs at whole-device granularity and retain synchronization barriers within rollout, leaving CPU and GPU capacity idle even with asynchronous rollout--training overlap.
To reclaim this capacity, we propose an efficient embodied RL training system \name.
\name combines an asynchronous pipelined scheduler with a fine-grained resource manager and provides programming support for users.
Evaluation across diverse models, simulators, and hardware shows that \name achieves $1.30\times$--$3.47\times$ the end-to-end trajectory throughput of the SOTA RL systems in each configuration.

\bibliographystyle{plain}
\bibliography{reference}

\clearpage
\appendix
\section{Greedy Core Rebinding}
\label{sec:appendix-core-rebinding}
Alg.~\ref{alg:greedy-step} shows the pseudocode for single-core workers.
When a core becomes free, the scheduler initializes an empty ready set (line~\#\ref{line:core-init}) and scans environments across all groups (lines~\#\ref{line:core-scan}--\ref{line:core-scan-end}).
For each environment, it checks that the current chunk has unfinished actions, no step is running, and the next action is available (line~\#\ref{line:core-ready}).
It adds eligible environments to the ready set (line~\#\ref{line:core-add}).
If the set is empty, the scheduler returns without dispatching work and leaves the core idle (lines~\#\ref{line:core-empty}--\ref{line:core-empty-end}).
% Role: mechanism -- follow selection, dispatch, and completion in order.
Else scheduler selects the ready environment with the largest $H[e]$ and atomically reserves it in the running set (line~\#\ref{line:core-select}).
It then rebinds the worker to the free core and dispatches one simulator step (line~\#\ref{line:core-dispatch}).
When the step completes, it advances the environment's action index and updates its measured latency (line~\#\ref{line:core-update}).
Finally, it removes the environment from the running set (line~\#\ref{line:core-clear}) and releases the core to trigger the next dispatch (line~\#\ref{line:core-release}).

\begin{algorithm}[h]
\footnotesize
\caption{Greedy core rebinding for single-core workers}
\label{alg:greedy-step}
\begin{algorithmic}[1]
\Require Free core $c$; environments $\mathcal{E}$; chunk length $K$; latency table $H$
\Ensure Dispatch one ready environment, if available
\State $\mathcal{E}_{\mathrm{ready}}\gets\emptyset$\label{line:core-init}
\ForAll{$e\in\mathcal{E}$}\label{line:core-scan}
  \If{$s_e<K\land e\notin\mathcal{E}_{\mathrm{running}}\land\textsc{Available}(a_{e,s_e})$}\label{line:core-ready}
    \State $\mathcal{E}_{\mathrm{ready}}\gets\mathcal{E}_{\mathrm{ready}}\cup\{e\}$\label{line:core-add}
  \EndIf
\EndFor\label{line:core-scan-end}
\If{$\mathcal{E}_{\mathrm{ready}}=\emptyset$}\label{line:core-empty}
  \State \Return
\EndIf\label{line:core-empty-end}
\State Atomically select $e^*\gets\arg\max_{e\in\mathcal{E}_{\mathrm{ready}}}H[e]$\label{line:core-select}
\Statex \hspace{\algorithmicindent}and add $e^*$ to $\mathcal{E}_{\mathrm{running}}$
\State Rebind worker $e^*$ to $c$; dispatch one step\label{line:core-dispatch}
\Statex \textit{On step completion:}
\State $s_{e^*}\gets s_{e^*}+1$; update $H[e^*]$\label{line:core-update}
\State $\mathcal{E}_{\mathrm{running}}\gets\mathcal{E}_{\mathrm{running}}\setminus\{e^*\}$\label{line:core-clear}
\State Release $c$ and trigger the next dispatch\label{line:core-release}
\end{algorithmic}
\end{algorithm}

\section{Simulation Performance}
\label{sec:appendix-simulation-performance}

\noindent\textbf{Simulation performance.}
Explicit CPU affinity improves simulator throughput and reduces latency.
Compared with default OS scheduling, our CPU bindings improve simulator throughput by factors of 1.38$\times$, 1.97$\times$, and 1.60$\times$ on LIBERO, RoboCasa, and BEHAVIOR, respectively (Table~\ref{tab:cpu_affinity_comparison}).
These benefits are due to CPU binding reducing scheduling preemption between environments.

\begin{table}[h]
\centering
\small
\caption{Simulator throughput with our optimal per-environment CPU bindings and default OS scheduling.}
\label{tab:cpu_affinity_comparison}
\begin{tabular}{@{}lcrr@{}}
\toprule
Benchmark & \shortstack{Cores/env} & \multicolumn{2}{c}{Throughput (steps/s)} \\
\cmidrule(l){3-4}
 & & Ours & OS default \\
\midrule
LIBERO & 1 & \textbf{1155.9} & 838.2 \\
RoboCasa & 1 & \textbf{700.90} & 355.9 \\
BEHAVIOR & 3 & \textbf{670.85} & 420.3 \\
\bottomrule
\end{tabular}
\end{table}

\end{document}